\documentclass[10pt,twocolumn,letterpaper]{article}

\usepackage{cvpr}
\usepackage{tikz}
\usepackage{times}
\usepackage{epsfig}
\usepackage{graphicx}
\usepackage{amsmath}
\usepackage{amssymb}
\usepackage{color}
\usepackage{booktabs}
\usepackage{xcolor}
\usepackage{balance}

\usetikzlibrary{backgrounds}
\usetikzlibrary{fadings}
\usetikzlibrary{patterns}
\usetikzlibrary{decorations}
\usetikzlibrary{decorations.pathmorphing}
\usetikzlibrary{decorations.pathreplacing}
\usetikzlibrary{arrows}
\usetikzlibrary{shapes}
\usetikzlibrary{calc}

\newcommand{\bp}{{\,\scriptscriptstyle\boxplus\,}}

\DeclareMathOperator*{\argmax}{argmax}

\newcommand{\quotes}[1]{``#1''}

\newcommand{\imgOverlay}[1]{
\begin{tikzpicture}
            \node[anchor=south east,inner sep=0] at (0,0) {\includegraphics[width=.098\linewidth]{images/tumMONO/img#1.jpg}};
            \node[anchor=south east,fill=white,font={\scriptsize},inner sep=0.4mm,fill opacity=0.6,text opacity=1] at (-0.3mm,0.3mm) {\color{black} #1};
\end{tikzpicture}}

\newcommand{\hl}[1]{#1}

\usepackage[pagebackref=true,breaklinks=true,colorlinks,bookmarks=true]{hyperref}

\cvprfinalcopy % *** Uncomment this line for the final submission

\def\cvprPaperID{****} % *** Enter the CVPR Paper ID here

\ifcvprfinal\fi
\begin{document}

%%%%%%%%% TITLE
\title{Direct Sparse Odometry}

\author{Jakob Engel\\
Technical University Munich\\
\and
Vladlen Koltun\\
Intel Labs\\
\and
Daniel Cremers\\
Technical University Munich{\color{white}\tiny\thanks{This work was supported by the ERC Consolidator Grant \quotes{3D Reloaded} and by a Google Faculty Research Award.}}\\
}
\maketitle
%\thispagestyle{empty}

%%%%%%%%% ABSTRACT
\begin{abstract}
We propose a novel direct sparse visual odometry formulation. 
It combines a fully direct probabilistic model (minimizing a photometric error) with 
consistent, joint optimization of all model parameters, including geometry 
-- represented as inverse depth in a reference frame -- and camera motion. 
This is achieved in real time by omitting the smoothness prior used in other 
direct methods and instead sampling pixels evenly throughout the images.
Since our method does not depend on keypoint detectors or descriptors, it can naturally 
sample pixels from across all image regions that have intensity gradient, including 
edges or smooth intensity variations on mostly white walls. The proposed model integrates a full photometric 
calibration, accounting for exposure time, lens vignetting, and non-linear 
response functions. 
We thoroughly evaluate our method on three different datasets comprising several hours of video. The experiments 
show that the presented approach significantly outperforms state-of-the-art 
direct and indirect methods in a variety of real-world settings, 
both in terms of tracking accuracy and robustness.
\end{abstract}

	\vspace{2mm}\section{Introduction}
	Simultaneous localization and mapping (SLAM) and visual odometry (VO)
	are fundamental building blocks for many emerging technologies -- from autonomous cars and UAVs 
	to virtual and augmented reality. 
	Realtime methods for SLAM and VO have made significant progress in recent
	years. While for a long time the field was dominated by feature-based (indirect) methods, in recent years a number of
	different approaches have gained in popularity, namely \textit{direct} and \textit{dense} formulations.
	
\begin{figure}
\centering
{\setlength{\fboxsep}{0pt}\setlength{\fboxrule}{0pt}\fbox{\includegraphics[width=.99\linewidth]{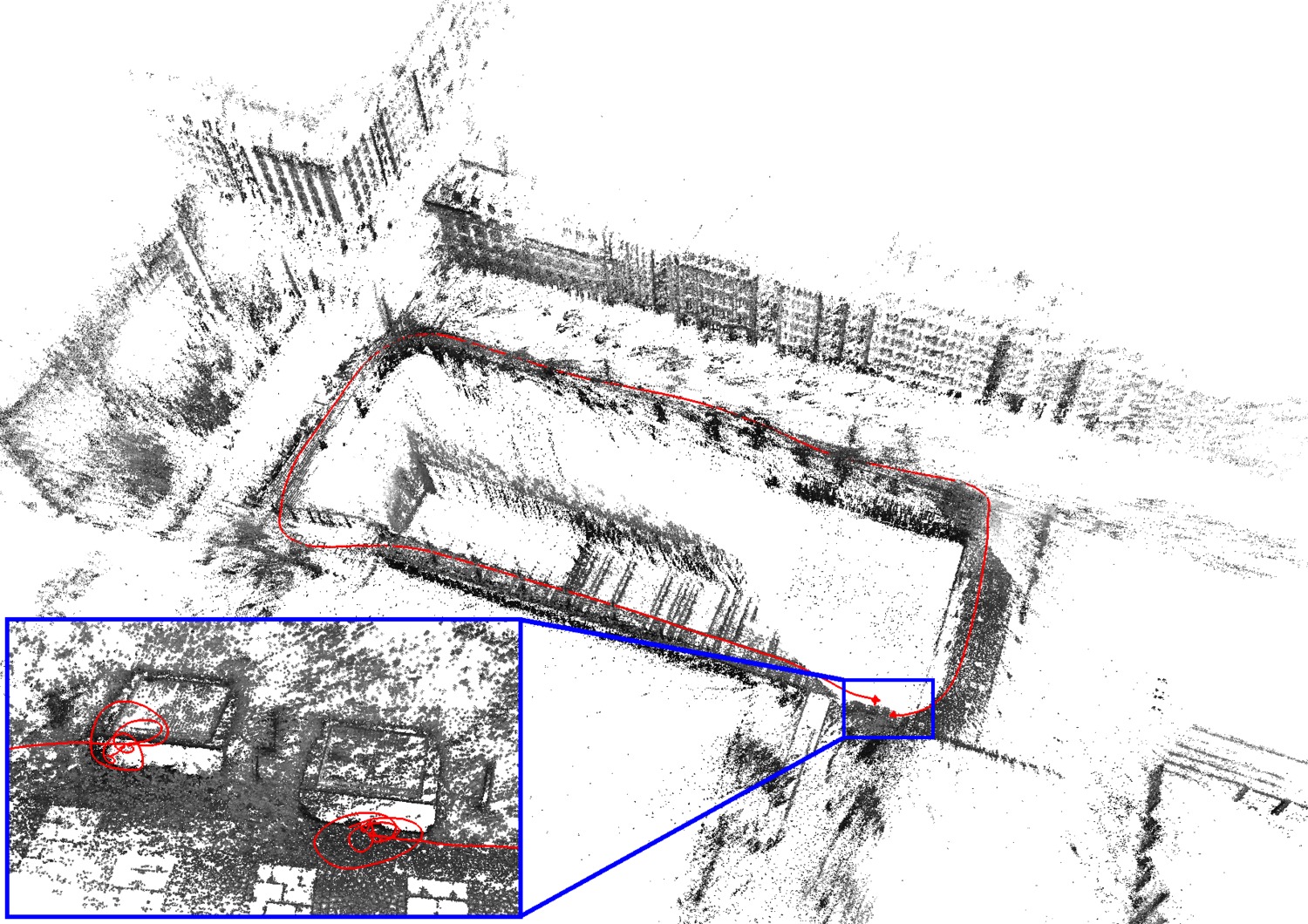}}}\\
\includegraphics[width=.195\linewidth]{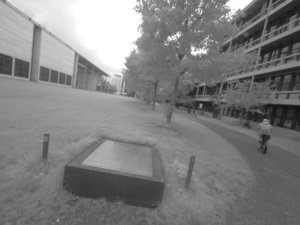}\hspace{-.5mm}
\includegraphics[width=.195\linewidth]{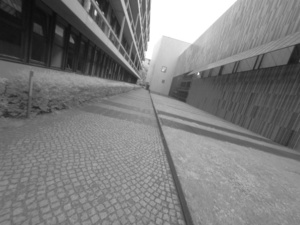}\hspace{-.5mm}
\includegraphics[width=.195\linewidth]{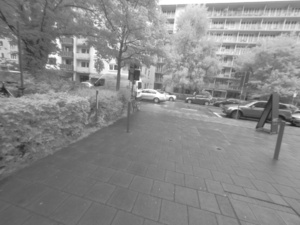}\hspace{-.5mm}
\includegraphics[width=.195\linewidth]{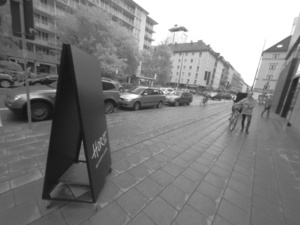}\hspace{-.5mm}
\includegraphics[width=.195\linewidth]{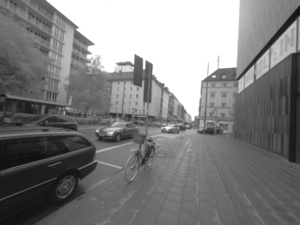}\\
\includegraphics[width=.195\linewidth]{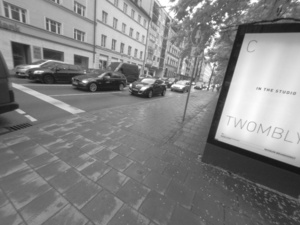}\hspace{-.5mm}
\includegraphics[width=.195\linewidth]{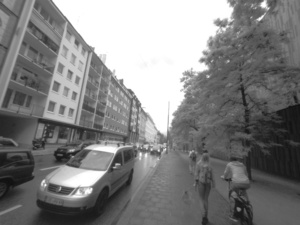}\hspace{-.5mm}
\includegraphics[width=.195\linewidth]{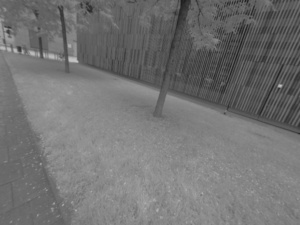}\hspace{-.5mm}
\includegraphics[width=.195\linewidth]{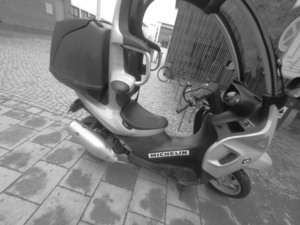}\hspace{-.5mm}
\includegraphics[width=.195\linewidth]{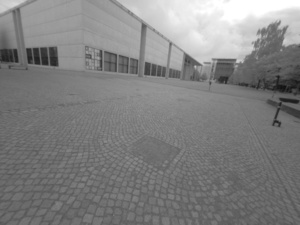}\\
\caption{\textbf{Direct sparse odometry (DSO).} 3D reconstruction and tracked trajectory for a 1:40min video cycling around a building (monocular visual odometry only). The bottom-left inset shows a close-up of the start and end point, 
visualizing the drift accumulated over the course of the trajectory. 
The bottom row shows some video frames.}
\label{fig:teaser}
\end{figure}

	\vspace{2mm}\paragraph{Direct vs. Indirect.}
	Underlying all formulations is a probabilistic model that takes noisy measurements $\mathbf{Y}$
	as input and computes an estimator $\mathbf{X}$ for the unknown, hidden model parameters (3D world model and camera motion).
	Typically a Maximum Likelihood approach is used, which finds the model parameters that maximize the probability 
	of obtaining the actual measurements, i.e., \mbox{$\mathbf{X}^* := \argmax_\mathbf{X} P(\mathbf{Y} | \mathbf{X})$}. 

	\textit{Indirect} methods then proceed in two steps. First, the raw sensor measurements are pre-processed to
	generate an intermediate representation, solving part of the overall problem, such as establishing correspondences. 
	Second, the computed intermediate values are
	interpreted as noisy measurements $\mathbf{Y}$ in a probabilistic model to estimate
	geometry and camera motion. 
	Note that the first step is typically approached by extracting and matching a sparse set of keypoints -- 
	however other options exist, like establishing correspondences in the form of dense, regularized optical flow.
	\hl{It can also include methods that extract and match parametric representations of other geometric primitives, 
	such as line- or curve-segments.}
	
	\textit{Direct} methods skip the pre-processing step and directly use the actual 
	sensor values -- light received from a certain direction over a certain time period -- as 
	measurements $\mathbf{Y}$ in a probabilistic model.
	
	In the case of passive vision, the direct approach thus optimizes a \textit{photometric error}, since the sensor
	provides photometric measurements. Indirect methods on the other hand optimize a \textit{geometric error},
	since the pre-computed values~-- point-positions or flow-vecors -- are geometric quantities.
	Note that for other sensor modalities like depth cameras or laser scanners (which directly measure 
	geometric quantities) direct formulations may also optimize a geometric error.

	\paragraph{Dense vs. Sparse.} 
	Sparse methods use and reconstruct only a selected set of independent points (traditionally corners), 
	whereas dense methods attempt to use and reconstruct all pixels in the 2D image domain. 
	Intermediate approaches (semi-dense) refrain from reconstructing the complete
	surface, but still aim at using and reconstructing a (largely connected and well-constrained) subset.
	
	Apart from the extent of the used image region however, a more fundamental -- and 
	consequential -- difference lies in the addition of a geometry prior.
	In the sparse formulation, there is no notion of neighborhood, and
	geometry parameters (keypoint positions) are conditionally independent given 
	the camera poses \& intrinsics\footnote{Note that even though early \hl{filtering-based methods \cite{jin00cvpr,davison07pami} kept} track 
	of point-point-correlations, these originated from marginalized camera poses, not from the model itself.}.
	Dense (or semi-dense) approaches on the other hand exploit the connectedness of the used image region to 
	formulate a geometry prior, typically favouring smoothness. In fact, such a prior is necessarily required to make 
	a dense world model observable from passive vision alone.
	In general, this prior is formulated directly in the form of an additional log-likelihood energy term \cite{stuehmer10dagm,newcombe2011iccv,pizzoli14icra}.\\
	
	Note that the distinction between \textit{dense and sparse} is not synonymous to \textit{direct and indirect} -- 
	in fact, all four combinations exist:
	\begin{itemize}
		\item \textbf{Sparse + Indirect:} This is the most widely-used formulation, estimating 3D geometry 
		from a set of keypoint-matches, thereby using a geometric error without a geometry prior.
		\hl{Examples include the work of Jin et al.~\cite{jin00cvpr},} monoSLAM \cite{davison07pami}, PTAM \cite{klein07ismar}, and ORB-SLAM \cite{mur2015orb}.
		\item \textbf{Dense + Indirect:} This formulation estimates 3D geometry from -- or in conjunction with -- a dense, 
		regularized optical flow field, thereby combining a geometric error (deviation from the flow field) with 
		a geometry prior (smoothness of the flow field), examples include \cite{valgaerts2012dense,ranftl16cvpr}.
		\item \textbf{Dense + Direct:} This formulation employs a photometric error as well as a geometric prior to estimate
		dense or semi-dense geometry. Examples include DTAM \cite{newcombe2011iccv}, its precursor \cite{stuehmer10dagm}, and LSD-SLAM \cite{engel14eccv}.
		\item \textbf{Sparse + Direct:} This is the formulation proposed in this paper. It optimizes a photometric error
		defined directly on the images, without incorporating a geometric prior. 
		\hl{While we are not aware of any recent work using this formulation, a sparse and direct formulation
		was already proposed by Jin et al.~in 2003 \cite{jin03js}.
		In contrast to their work however, which is based on an extended Kalman filter, our method 
		uses a non-linear optimization framework. The motivation for exploring the combination of sparse and direct 
		is laid out in the following section.}
	\end{itemize}

\subsection{Motivation}
	The \textbf{direct and sparse} formulation for monocular visual odometry proposed in this paper is motivated by the following considerations.

	\textbf{(1) Direct:} One of the main benefits of keypoints is their ability to provide robustness to 
	photometric and geometric distortions present in images taken with off-the-shelf commodity cameras. Examples are
	automatic exposure changes, non-linear response functions (gamma correction / white-balancing), lens attenuation 
	(vignetting), de-bayering artefacts, or even strong geometric distortions caused by a rolling shutter. 
	
	At the same time, for all use-cases mentioned in the introduction, millions of devices will be (and already are) 
	equipped with cameras solely meant to provide data for computer vision algorithms, 
	instead of capturing images for human consumption. These cameras 	
	should and will be designed to provide a complete sensor model, and to
	capture data in a way that best serves the processing algorithms:
	Auto-exposure and gamma correction for instance 
	are not unknown noise sources, but features that provide better image data -- and that can 
	be incorporated into the model, making the obtained data more informative. 
	Since the direct approach models the full image formation process down to pixel intensities, it
	greatly benefits from a more precise sensor model. 
	
	One of the main benefits of a direct formulation is that it does 
	not require a point to be recognizable by itself, thereby allowing
	for a more finely grained geometry representation (pixelwise inverse depth). Furthermore, we can sample
	from across all available data -- including edges and weak intensity variations -- generating a more 
	complete model and lending more robustness in sparsely textured environments.

\begin{figure}
\centering
\includegraphics[width=.49\linewidth]{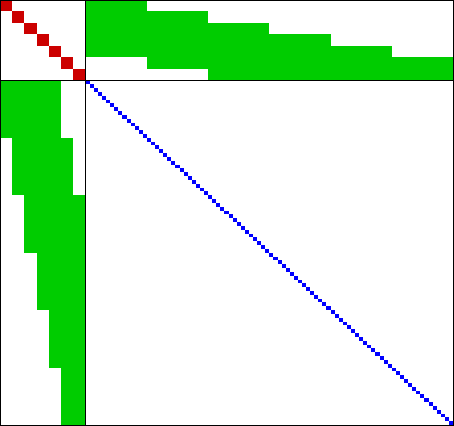} \hfill
\includegraphics[width=.49\linewidth]{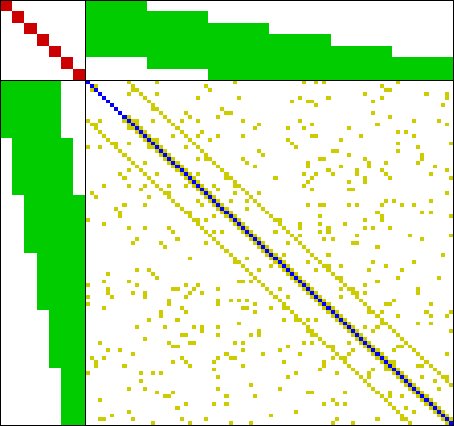}\\[-0.1cm]
{\setlength{\fboxsep}{1pt}\setlength{\fboxrule}{0.5pt}\fbox{\scriptsize
\includegraphics[height=0.2cm]{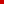} pose (diag) \hspace{0.17cm}
\includegraphics[height=0.2cm]{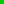} pose-geo \hspace{0.17cm}
\includegraphics[height=0.2cm]{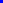} geo (diag) \hspace{0.17cm}
\includegraphics[height=0.2cm]{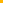} geo (off-diag)}}\\[2mm]
\caption{\textbf{Sparse vs.~dense Hessian structure.} 
Left: Hessian structure of sparse bundle adjustment: since the geometry-geometry block is diagonal, 
it can be solved efficiently using the Schur complement. Right: A geometry prior adds (partially unstructured) geometry-geometry
correlations -- the resulting system is hence not only much larger, but also becomes much harder to solve. 
For simplicity, we do not show the global camera intrinsic parameters.}
\label{fig:hessian}
\end{figure}

	\textbf{(2) Sparse:} 
	The main drawback of adding a geometry prior is the introduction of correlations between geometry parameters, which
	render a statistically consistent, joint optimization in real time 
	infeasible (see Figure~\ref{fig:hessian}). This is why existing dense or semi-dense approaches (a) neglect or 
	coarsely approximate correlations between geometry parameters (orange), and / or between geometry parameters and 
	camera poses (green), and (b) employ different optimization methods for the dense geometry part, such as a primal-dual 
	formulation \cite{stuehmer10dagm,newcombe2011iccv,pizzoli14icra}.
	
	In addition, the expressive complexity of today's priors is limited: 
%	They often simply 
%	encourage a smooth world, are restricted to a few
%	object classes like cars \cite{guney2015cvpr}, chairs and tables \cite{salas2013cvpr}, 
%	or planes \cite{salas2014ismar}\footnote{Strictly 
%	speaking, \cite{guney2015cvpr,salas2013cvpr,salas2014ismar} all do not impose a probabilistic prior, 
%	but rather intentionally limit the models expressive power, by replacing parts of it with a fixed 
%	3D model that has less degrees of freedom}, or operate in an offline, batch setup \cite{hane2013cvpr}.
	While they make the 3D reconstruction denser, locally more accurate and more visually appealing, 
	we found that priors can introduce a bias, and thereby reduce rather than increase long-term, large-scale accuracy. 
	Note that in time this may well change with the introduction of more realistic, unbiased priors learnt from real-world data.\\
	
	\subsection{Contribution and Outline}
	In this paper we propose a sparse and direct approach to monocular visual odometry. To our knowledge,
	it is the only fully direct method that jointly optimizes the full likelihood for all involved model parameters, including 
	camera poses, camera intrinsics, and geometry parameters (inverse depth values). 
	This is in contrast to hybrid approaches such as SVO \cite{forster14icra}, which revert 
	to an indirect formulation for joint model optimization.
	
	Optimization is performed in a sliding window, where old camera poses as well as points that leave the field
	of view of the camera are marginalized, in a manner inspired by \cite{leutenegger2015ijrr}. 
	%However, since the proposed 
	%direct sparse model has the same structure as indirect methods, other back-end approaches like 
	%filtering [MSCKF], Incremental Smoothing and Mapping [iSAM2] or full bundle adjustment can also be used,
	%as long as sufficiently accurate initializations can be obtained.
	In contrast to existing approaches, our method further takes full advantage of photometric 
	camera calibration, including lens attenuation, gamma correction, 
	and known exposure times. This integrated photometric calibration further increases accuracy and robustness.
	
	Our CPU-based implementation runs in real time on a laptop computer.
	We show in extensive evaluations on three different datasets comprising several hours of video that it 
	outperforms other state-of-the-art approaches (direct and indirect), both in terms of robustness and accuracy.
	With reduced settings (less points and active keyframes), it even runs at $5\!\times $ real-time 
	speed while still outperforming state-of-the-art indirect methods. On high, non-real-time settings in turn (more 
	points and active keyframes), it creates semi-dense models similar in density to those of LSD-SLAM, but much more accurate. \\

	The paper is organized as follows: The proposed direct, sparse model as well as the windowed optimization method are 
	described in Section~\ref{secDSVO}. Specifically, this comprises the geometric and photometric camera calibration in 
	Section~\ref{ssecCalibration}, the model formulation in Section~\ref{ssecModel}, and the windowed optimization in Section~\ref{ssecOptimization}.
	Section \ref{secFrontEnd} describes the front-end: the part of the algorithm that performs data-selection and provides
	sufficiently accurate initializations for the highly non-convex optimization back-end.
	We provide a thorough experimental comparison to other methods in Section~\ref{ssecQuantComp}. We also evaluate the effect of important
	parameters and new concepts like the use of photometric calibration in Section~\ref{ssecParameter}. In Section~\ref{ssecResultsNoise}, 
	we analyse the effect of added photometric and geometric noise to the data.
	Finally, we provide a summary in Section~\ref{secConclusion}.

	\section{Direct Sparse Model}
	\label{secDSVO}
	Our direct sparse odometry is based on continuous optimization
	of the photometric error over a window of recent frames, taking into account a photometrically calibrated 
	model for image formation. In contrast to existing direct methods, 
	we jointly optimize for all involved parameters (camera intrinsics, camera extrinsics, and inverse depth values), 
	effectively performing the photometric equivalent of windowed sparse bundle adjustment.	
	We keep the geometry representation employed by other direct approaches, i.e.,
	3D points are represented as inverse depth in a reference frame (and thus have one degree of freedom).\\
		
	\paragraph{Notation.}
	Throughout the paper, bold lower-case letters ($\mathbf{x}$) represent vectors and bold
	upper-case letters ($\mathbf{H}$) represent matrices. Scalars will be represented by light lower-case letters ($t$),
	functions (including images) by light upper-case letters ($I$).
	Camera poses are represented as transformation matrices $\mathbf{T}_i \in \text{SE(3)}$, transforming
	a point from the world frame into the camera frame. Linearized pose-increments will be expressed as
	Lie-algebra elements $\boldsymbol{x}_i \in \mathfrak{se}\text{(3)}$, which -- with a slight abuse of notation -- we 
	directly write as vectors $\boldsymbol{x}_i \in \mathbb{R}^6$.
	We further define the commonly used operator $\bp\colon \mathfrak{se}\text{(3)} \times \text{SE(3)} \to \text{SE(3)}$ 
	using a left-multiplicative formulation, i.e., 
	\begin{align}
		\label{eq:boxplus}
		\boldsymbol{x}_i \bp \mathbf{T}_i := e^{\widehat{\boldsymbol{x}_i}} \cdot \mathbf{T}_i.
	\end{align}

	\subsection{Calibration}
	\label{ssecCalibration}
	The direct approach comprehensively models the image formation process.
	In addition to a \textit{geometric} camera model -- which comprises the function that projects a 
	3D point onto the 2D image -- it is hence beneficial to also consider a \textit{photometric} camera model, 
	which comprises the function that maps real-world energy received by a pixel on the sensor (irradiance) to
	the respective intensity value. 
	Note that for indirect methods this is of little benefit and hence widely ignored, as common feature extractors and 
	descriptors are invariant (or highly robust) to photometric variations.

\begin{figure}
\centering
\includegraphics[width=.99\linewidth]{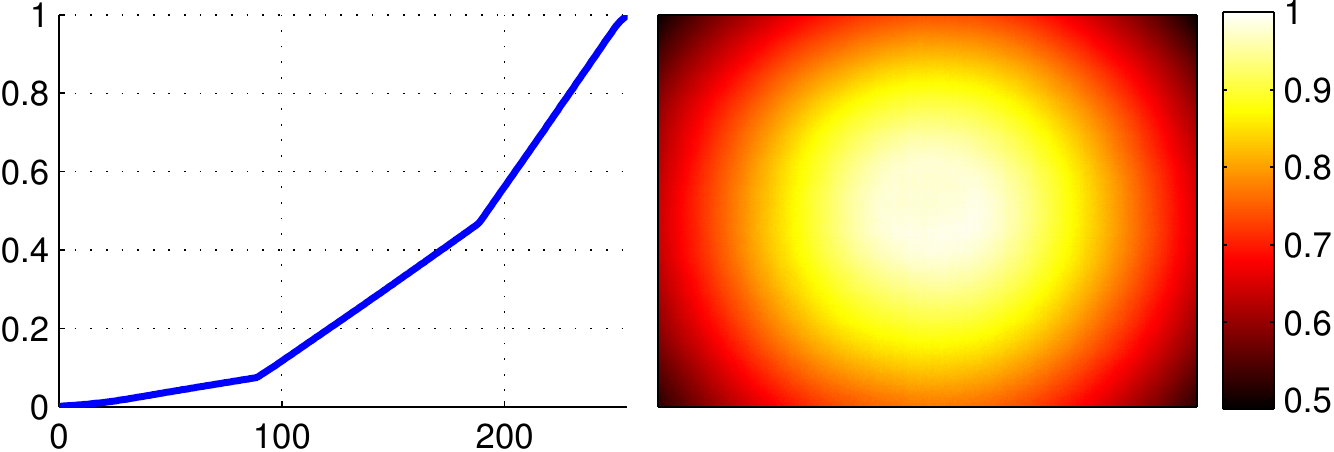}\\
\includegraphics[width=.99\linewidth]{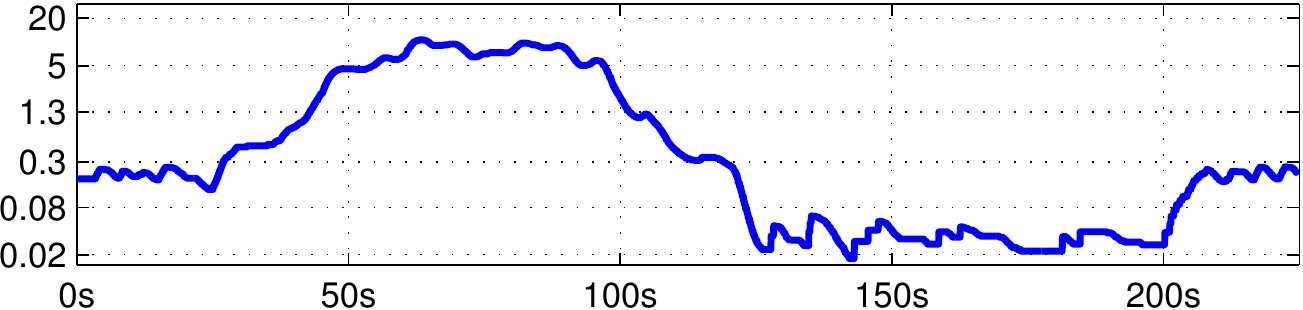}\\
\caption{\textbf{Photometric calibration.} 
Top: Inverse response function $G^{-1}$ and lens attenuation $V$ of the camera used for Figure~\ref{fig:teaser}. 
Bottom: Exposure $t$ in milliseconds for a sequence containing an indoor and an outdoor part.
Note how it varies by a factor of more than 500, from 0.018 to 10.5ms. 
Instead of treating these quantities as unknown noise sources, we explicitly 
account for them in the photometric error model.}
\label{fig:camCalib}
\end{figure}

	\subsubsection{Geometric Camera Calibration}
	For simplicity, we formulate our method for the well-known pinhole camera model -- radial distortion is removed
	in a preprocessing step. While for wide-angle cameras this does reduce 
	the field of view, it allows comparison across methods that only implement a limited choice of camera models. 
	Throughout this paper, we will denote projection by $\Pi_\mathbf{c} \colon \mathbb{R}^3 \to \Omega$
	and back-projection with $\Pi_\mathbf{c}^{-1} \colon \Omega \times \mathbb{R} \to \mathbb{R}^3$,
	where $\mathbf{c}$ denotes the intrinsic camera parameters (for the pinhole model these are the focal length and the principal point).
	Note that analogously to \cite{caruso2015iros}, our approach can be extended to other (invertible) camera models,
	although this does increase computational demands.

	\subsubsection{Photometric Camera Calibration}
	We use the image formation model used in \cite{engel16archiveDataset}, which accounts for a non-linear response 
	function $G \colon \mathbb{R} \to [0,255]$, as well as lens attenuation (vignetting) $V \colon \Omega \to [0,1]$.
	Figure~\ref{fig:camCalib} shows an example calibration from the TUM monoVO dataset.
	The combined model is then given by
	\begin{align}
		I_i(\mathbf{x}) = G\big(t_i V(\mathbf{x}) B_i(\mathbf{x})\big),
	\end{align}
	where $B_i$ and $I_i$ are the irradiance and the observed pixel intensity in frame $i$, and $t_i$ is the exposure time.
	The model is applied by photometrically correcting each video frame as very first step, by computing
	\begin{align}
		I'_i(\mathbf{x}) := t_i B_i(\mathbf{x}) = \frac{G^{-1}(I_i(\mathbf{x}))}{V(\mathbf{x})}.
	\end{align}
	In the remainder of this paper, $I_i$ will always refer to the photometrically corrected image $I'_i$, except where otherwise stated.

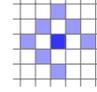
\begin{figure}
\centering
	\begin{tikzpicture}
	\fill[blue!80!white] (0.4,0.4) rectangle (0.6,0.6);
	\fill[blue!40!white] (0.4,0.4)++(0.4,0) rectangle +(0.2,0.2);
	\fill[blue!40!white] (0.4,0.4)++(-0.4,0) rectangle +(0.2,0.2);
	\fill[blue!40!white] (0.4,0.4)++(0,0.4) rectangle +(0.2,0.2);
	\fill[blue!40!white] (0.4,0.4)++(0,-0.4) rectangle +(0.2,0.2);
	\fill[blue!40!white] (0.4,0.4)++(0.2,0.2) rectangle +(0.2,0.2);
	\fill[blue!40!white] (0.4,0.4)++(-0.2,0.2) rectangle +(0.2,0.2);
	\fill[blue!40!white] (0.4,0.4)++(-0.2,-0.2) rectangle +(0.2,0.2);
	\draw[step=.2cm,gray,thin] (-0.1,-0.1) grid (1.1,1.1);	
	\end{tikzpicture}\\
\caption{\textbf{Residual pattern.} Pattern $\mathcal{N}_\mathbf{p}$ used for energy computation. The bottom-right pixel is omitted to enable 
SSE-optimized processing. Note that since we have 1 unknown per point (its inverse depth),
and do not use a regularizer, we require $|\mathcal{N}_\mathbf{p}| > 1$ in order for all model parameters
to be well-constrained when optimizing over only two frames.
Figure~\ref{fig:patternEval} shows an evaluation of how this pattern affects tracking accuracy.}
\label{fig:pattern}
\end{figure}

	\subsection{Model Formulation}
	\label{ssecModel}
	We define the photometric error of a point $\mathbf{p} \in \Omega_i$ in reference frame $I_i$, 
	observed in a target frame $I_j$, as the weighted
	SSD over a small neighborhood of pixels. Our experiments have shown that 8 pixels, arranged in a slightly spread pattern
	(see Figure~\ref{fig:pattern}) give a good trade-off between computations required for evaluation, robustness to motion blur, and providing 
	sufficient information. 
	Note that in terms of the contained information, evaluating the SSD over such a small neighborhood of pixels is similar to 
	adding first- and second-order irradiance derivative constancy terms (in addition to irradiance constancy) for the central pixel.
	Let
	\begin{align}
		\label{eq:PixelPhotErr}
		E_{\mathbf{p}j} &:= 
			\sum_{\mathbf{p} \in \mathcal{N}_\mathbf{p}} w_\mathbf{p} \left\|(I_j[\mathbf{p}']\!-\!b_j) - \frac{t_je^{a_j}}{t_ie^{a_i}} \bigl(I_i[\mathbf{p}]\!-\!b_i\bigr)\right\|_\gamma,
	\end{align}
	where $\mathcal{N}_\mathbf{p}$ is the set of pixels included in the SSD; $t_i, t_j$ the exposure 
	times of the images $I_i, I_j$; and $\|\cdot\|_\gamma$ the Huber norm. Further, $\mathbf{p}'$ stands for the projected point position of $\mathbf{p}$ with
	inverse depth $d_\mathbf{p}$, given by
	\begin{align}
	\mathbf{p}' = \Pi_\mathbf{c}\big(\mathbf{R} ~\Pi_\mathbf{c}^{-1}(\mathbf{p}, d_\mathbf{p}) + \mathbf{t}\big),
	\end{align}
	with
	\begin{align}
	\begin{bmatrix}\mathbf{R}&\mathbf{t}\\0&1\end{bmatrix} := \mathbf{T}_j \mathbf{T}_i^{-1}.
	\end{align}
	In order to allow our method to operate on sequences without known exposure times, we include an 
	additional affine brightness transfer function given by $e^{-a_i} (I_i- b_i)$. 
	Note that in contrast to most previous formulations \cite{jin03js,engel15iros}, the scalar factor 
	$e^{-a_i}$ is parametrized logarithmically. 
	This both prevents it from becoming negative, and avoids numerical issues arising from 
	multiplicative (i.e., exponentially increasing) drift.
	
	In addition to using robust Huber penalties, we apply a gradient-dependent weighting $w_\mathbf{p}$ given by
	\begin{align}
		w_\mathbf{p} &:= \frac{c^2}{c^2 + \|\nabla I_i(\mathbf{p})\|_2^2},
	\end{align}
	which down-weights pixels with high gradient. This weighting function can be probabilistically interpreted as adding
	small, independent geometric noise on the projected point position $\mathbf{p}'$, and immediately marginalizing 
	it -- approximating small geometric error. 
	To summarize, the error $E_{\mathbf{p}j}$ depends on the following variables: (1) the point's inverse depth $d_\mathbf{p}$, (2) the camera intrinsics $\mathbf{c}$, (3) the poses of the involved frames $\mathbf{T}_i, \mathbf{T}_j$, and (4) their brightness transfer function parameters $a_i, b_i, a_j, b_j$.
		
\begin{figure}
\centering

	\begin{tikzpicture}

	\draw (-3cm, 4.5cm) node[shade, bottom color=blue!10, top color=blue!2,draw,ellipse, minimum width=0.8cm, minimum height=0.9cm, inner sep=0mm] (KF1) 
	{\begin{minipage}{1.5cm}\centering \scriptsize KF 1: \\$\mathbf{T}_1, a_1, b_1$\end{minipage}};
	\draw (-3cm, 3cm) node[shade, bottom color=blue!10, top color=blue!2,draw,ellipse, minimum width=0.8cm, minimum height=0.9cm, inner sep=0mm] (KF2) 
	{\begin{minipage}{1.5cm}\centering \scriptsize KF 2: \\$\mathbf{T}_2, a_2, b_2$\end{minipage}};
	\draw (-3cm, 2.0cm) node[shade, bottom color=blue!10, top color=blue!2,draw,ellipse, minimum width=0.8cm, minimum height=0.9cm, inner sep=0mm] (KF3) 
	{\begin{minipage}{1.5cm}\centering \scriptsize KF 3: \\$\mathbf{T}_3, a_3, b_3$\end{minipage}};
	\draw (-3cm, 1.0cm) node[shade, bottom color=blue!10, top color=blue!2,draw,ellipse, minimum width=0.8cm, minimum height=0.9cm, inner sep=0mm] (KF4)
	{\begin{minipage}{1.5cm}\centering \scriptsize KF 4: \\$\mathbf{T}_4, a_4, b_4$\end{minipage}};

	\draw (3cm, 4.5cm) node[shade, bottom color=blue!10, top color=blue!2,draw,ellipse, inner sep=0.5mm] (d1) {\scriptsize Pt 1: $d_\mathbf{p_1}$};
	
	\draw (3cm, 3.3cm) node[shade, bottom color=blue!10, top color=blue!2,draw,ellipse, inner sep=0.5mm] (d2) {\scriptsize Pt 2: $d_\mathbf{p_2}$};
	\draw (3cm, 2.7cm) node[shade, bottom color=blue!10, top color=blue!2,draw,ellipse, inner sep=0.5mm] (d3) {\scriptsize Pt 3: $d_\mathbf{p_3}$};
	
	\draw (3cm, 1cm) node[shade, bottom color=blue!10, top color=blue!2,draw,ellipse, inner sep=0.5mm] (d4) {\scriptsize Pt 4: $d_\mathbf{p_4}$};

	\draw (1cm, 4.7cm) node[shade, bottom color=blue!10, top color=blue!2,draw,rectangle, inner sep=0.5mm] (r12) {\tiny $E_{\mathbf{p}_12}$};
	\draw (r12.east) -- (d1.west);
	\draw[blue!80!black] (r12.west) -- (KF1.east);
	\draw[red!80!black] (r12.west) -- (KF2.east);
	
	\draw (1cm, 4.3cm) node[shade, bottom color=blue!10, top color=blue!2,draw,rectangle, inner sep=0.5mm] (r13)  {\tiny $E_{\mathbf{p}_13}$};
	\draw (r13.east) -- (d1.west);
	\draw[blue!80!black] (r13.west) -- (KF1.east);
	\draw[red!80!black] (r13.west) -- (KF3.east);

	\draw (1cm, 3.6cm) node[shade, bottom color=blue!10, top color=blue!2,draw,rectangle, inner sep=0.5mm] (r21)  {\tiny $E_{\mathbf{p}_21}$};
	\draw (r21.east) -- (d2.west);
	\draw[blue!80!black] (r21.west) -- (KF2.east);
	\draw[red!80!black] (r21.west) -- (KF1.east);
	\draw (1cm, 3.2cm) node[shade, bottom color=blue!10, top color=blue!2,draw,rectangle, inner sep=0.5mm] (r23)  {\tiny $E_{\mathbf{p}_23}$};
	\draw (r23.east) -- (d2.west);
	\draw[blue!80!black] (r23.west) -- (KF2.east);
	\draw[red!80!black] (r23.west) -- (KF3.east);
	\draw (1cm, 2.8cm) node[shade, bottom color=blue!10, top color=blue!2,draw,rectangle, inner sep=0.5mm] (r33)  {\tiny $E_{\mathbf{p}_33}$};
	\draw (r33.east) -- (d3.west);
	\draw[blue!80!black] (r33.west) -- (KF2.east);
	\draw[red!80!black] (r33.west) -- (KF3.east);
	\draw (1cm, 2.4cm) node[shade, bottom color=blue!10, top color=blue!2,draw,rectangle, inner sep=0.5mm] (r34)  {\tiny $E_{\mathbf{p}_34}$};
	\draw (r34.east) -- (d3.west);
	\draw[blue!80!black] (r34.west) -- (KF2.east);
	\draw[red!80!black] (r34.west) -- (KF4.east);

	\draw (1cm, 1.4cm) node[shade, bottom color=blue!10, top color=blue!2,draw,rectangle, inner sep=0.5mm] (r41)  {\tiny $E_{\mathbf{p}_41}$};
	\draw (r41.east) -- (d4.west);
	\draw[blue!80!black] (r41.west) -- (KF4.east);
	\draw[red!80!black] (r41.west) -- (KF1.east);
	
	\draw (1cm, 1.0cm) node[shade, bottom color=blue!10, top color=blue!2,draw,rectangle, inner sep=0.5mm] (r42)  {\tiny $E_{\mathbf{p}_42}$};
	\draw (r42.east) -- (d4.west);
	\draw[blue!80!black] (r42.west) -- (KF4.east);
	\draw[red!80!black] (r42.west) -- (KF2.east);
	
	\draw (1cm, 0.6cm) node[shade, bottom color=blue!10, top color=blue!2,draw,rectangle, inner sep=0.5mm] (r43)  {\tiny $E_{\mathbf{p}_43}$};
	\draw (r43.east) -- (d4.west);
	\draw[blue!80!black] (r43.west) -- (KF4.east);
	\draw[red!80!black] (r43.west) -- (KF3.east);

	\end{tikzpicture}\\[1mm]
	\caption{\textbf{Factor graph for the direct sparse model.} Example with four keyframes and four points; one in KF1, two in KF2, and one in KF4. 
	Each energy term (defined in Eq. (\ref{eq:PixelPhotErr})) depends on the point's host frame (blue), the frame the point is observed in (red), 
	and the point's inverse depth (black). Further, all terms depend on the global camera intrinsics vector $\mathbf{c}$, which is not shown.}
	\label{fig:factorgraph}
	\vspace{-5mm}
\end{figure}
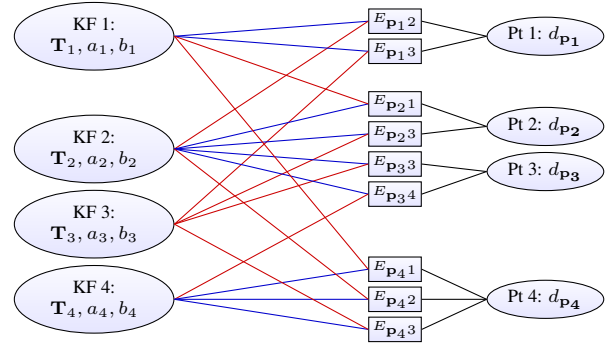

	The full photometric error over all frames and points is given by
	\begin{align}
		\label{eq:FullPhotErr}
		E_\text{photo} := \sum_{i\in\mathcal{F}} \sum_{\mathbf{p}\in\mathcal{P}_i} \sum_{j\in\text{obs}(\mathbf{p})} E_{\mathbf{p}j}.
	\end{align}
	where $i$ runs over all frames $\mathcal{F}$, $\mathbf{p}$ over all points $\mathcal{P}_i$ in frame $i$, and $j$ over
	all frames $\text{obs}(\mathbf{p})$ in which the point $\mathbf{p}$ is visible. Figure~\ref{fig:factorgraph} shows the 
	resulting factor graph: The only difference to the classical reprojection error is the additional dependency of each residual
	on the pose of the host frame, i.e., each term depends on \textit{two} frames instead of only one. While this adds
	off-diagonal entries to the pose-pose block of the Hessian, it does not affect the sparsity pattern \textit{after} application of 
	the Schur complement to marginalize point parameters. The resulting system can thus be solved analogously to the indirect formulation.
	Note that the Jacobians with respect to the two frames' poses are linearly related by the adjoint 
	of their relative pose. In practice, this factor can then be pulled out of the sum when computing the Hessian or its
	Schur complement, greatly reducing the additional computations caused by more variable dependencies.
	
	If exposure times are known, we further add a prior pulling the affine brightness transfer function to zero:
	\begin{align}
	 E_\text{prior} := \sum_{i\in\mathcal{F}} \left(\lambda_a a_i^2 + \lambda_b b_i^2\right).
	\end{align}
	If no photometric calibration is available, we set $t_i=1$ and $\lambda_a = \lambda_b = 0$, as in this case they need to 
	model the (unknown) changing exposure time of the camera. 
	As a side-note it should be mentioned that the ML estimator for a multiplicative factor 
	$a^* = \argmax_a \sum_i(ax_i-y_i)^2$ is biased if both $x_i$ and $y_i$ contain noisy measurements (see \cite{engel14ras});
	causing $a$ to drift in the unconstrained case $\lambda_a=0$. While this generally has little 
	effect on the estimated poses, it may introduce a bias if the scene contains only few, weak intensity variations.

	\paragraph{Point Dimensionality.}
	In the proposed direct model, a point is parametrized by only one parameter 
	(the inverse depth in the reference frame), in contrast to three unknowns as in the indirect model. 
	To understand the reason for this difference, we first note that in both cases a 3D point is in fact an 
	arbitrarily located discrete sample on a continuous, real-world 3D surface.
	The difference then lies in the way this 2D location on the surface is defined.
	In the indirect approach, it is implicitly defined as \textit{the point, which (projected 
	into an image) generates a maximum in the used corner response function}. This entails that both the 
	surface, as well as the point's location on the surface are unknowns, and need to be estimated.
	In our direct formulation, a point is simply defined as \textit{the point, where the 
	source pixel's ray hits the surface}, thus only one unknown remains.
	In addition to a reduced number of parameters, this naturally enables an \textit{inverse depth} parametrization,
	which -- in a Gaussian framework -- is better suited to represent uncertainty from stereo-based depth estimation,
	in particular for far-away points \cite{civera_inverse_2008}.

	\paragraph{Consistency.} Strictly speaking, the proposed direct sparse model does allow to use 
	some observations (pixel values) multiple times, while others are not used at all. 
	This is because -- even though our point selection strategy attempts 
	to avoid this by equally distributing points in space (see Section~\ref{sssecpointManagement}) -- we 
	allow point observations to overlap, and thus depend on 
	the same pixel value(s). This particularly happens in scenes with little texture, where all
	points have to be chosen from a small subset of textured image regions.
	We however argue that this has negligible effect in practice, and -- if desired -- can be avoided by removing (or downweighting) observations that use the same pixel value.

	\subsection{Windowed Optimization}
	\label{ssecOptimization}
	We follow the approach by Leutenegger et al.~\cite{leutenegger2015ijrr} and optimize the total error (\ref{eq:FullPhotErr}) 
	in a sliding window using the Gauss-Newton algorithm, which gives a good trade-off between speed and flexibility.
%	Note that the proposed direct framework is also compatible with other state-of-the-art back-end approaches like EKF-based 
%	filtering \cite{murikis}, Asynchronous Adaptive Conditioning \cite{Sibley} or incremental Smoothing and Mapping \cite{kaessEtAl},
%	which can be used to extend it to full SLAM including loop-closures, or add integration of inertial data. 
	
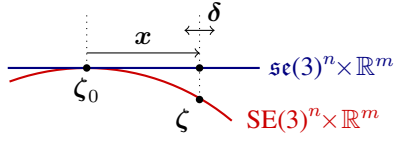
\begin{figure}
\centering
	\begin{tikzpicture}
	\draw [thick,domain=110:50,red!80!black] plot ({3*cos(\x)}, {3*sin(\x)});
	\draw (2,2.3)  node[anchor=west,red!80!black] {$\text{SE(3)}^n\!\!\times\!\mathbb{R}^m$};
	\draw [thick,blue!50!black] (-1.05,3) -- (2.3,3) node[pos=1, anchor=west] {$\mathfrak{se}\text{(3)}^n\!\!\times\!\mathbb{R}^m$};
	
	\draw (0,3) node[circle,fill,black,scale=0.3] {} node[anchor=north] {$\boldsymbol{\zeta}_0$};
	\draw (1.5,3) node[circle,fill,black,scale=0.3] {};
	\draw (1.5,2.58) node[circle,fill,black,scale=0.3] {} node[anchor=north east] {$\boldsymbol{\zeta}$};
	\draw [dotted] (1.5,3.7) -- (1.5,2.58);
	\draw [dotted] (0,3.7) -- (0,3);
	\draw[black,->] (0,3.2) -- (1.5,3.2) node[anchor=south,pos=0.5] {$\boldsymbol{x}$};
	\draw[black,<->] (1.3,3.5) -- (1.7,3.5) node[anchor=south,pos=1] {$\boldsymbol\delta$};

	\end{tikzpicture}
	\caption{\textbf{Windowed optimization.} The red curve denotes the parameter space, composed of non-Euclidean camera poses in $\text{SE}(3)$, and 
	the remaining Euclidean parameters. The blue line corresponds to the
	tangent-space around $\boldsymbol{\zeta}_0$, in which we (1) accumulate the quadratic marginalization-prior on $\boldsymbol{x}$, 
	and (2) compute Gauss-Newton steps $\boldsymbol\delta$. 
	For each parameter, the tangent space is fixed as soon as that parameter becomes part of the marginalization term. 
	Note that while we treat all parameters equally in our notation, for Euclidean parameters tangent-space and parameter-space coincide.}
	\label{fig:manifold}
	\vspace{-5mm}
\end{figure}

	For ease of notation, we extend the $\bp$ operator as defined in (\ref{eq:boxplus}) to all optimized parameters -- 
	for parameters other than $\text{SE(3)}$ poses it denotes conventional addition. We will use $\boldsymbol{\zeta} \in \text{SE(3)}^n\!\times\!\mathbb{R}^m$ to denote all optimized variables,
	including camera poses, affine brightness parameters, inverse depth values, and camera intrinsics.
	As in \cite{leutenegger2015ijrr}, marginalizing a residual that depends on a parameter in $\boldsymbol{\zeta}$ will fix the tangent space in 
	which any future information (delta-updates) on that parameter is accumulated. 
	We will denote the evaluation point for this tangent space with $\boldsymbol{\zeta}_0$, 
	and the accumulated delta-updates by $\boldsymbol{x} \in \mathfrak{se}\text{(3)}^n\!\times\!\mathbb{R}^m$. The current 
	state estimate is hence given by $\boldsymbol{\zeta} = \boldsymbol{x} \bp \boldsymbol{\zeta}_0$.
	Figure~\ref{fig:manifold} visualizes the relation between the different variables. 
	
	\paragraph{Gauss-Newton Optimization.}
	We compute the Gauss-Newton system as 
	\begin{align}
		\label{eq:GNApprox}
		\mathbf{H} = \mathbf{J}^T \mathbf{W} \mathbf{J} \text{~~~~ and ~~~~} \mathbf{b} = -\mathbf{J}^T \mathbf{W} \mathbf{r},
	\end{align}
	where $\mathbf{W} \in \mathbb{R}^{n \times n}$ is the diagonal matrix containing the weights, $\mathbf{r} \in \mathbb{R}^n$ 
	is the stacked residual vector, and $\mathbf{J} \in \mathbb{R}^{n \times d}$ is the Jacobian of $\mathbf{r}$. 
	
	Note that each point contributes $|\mathcal{N}_p|=8$ residuals to the energy. For notational simplicity, we will in the following
	consider only a single residual $r_k$, and the associated row of the Jacobian $\mathbf{J}_k$. 
	During optimization -- as well as when marginalizing -- residuals are always evaluated at the current state estimate, i.e.,
	\begin{align}
		\label{eq:JacR}
		r_k &= r_k(\boldsymbol{x} \bp \boldsymbol{\zeta}_0)\\
	\nonumber
		&= \big(I_j[\mathbf{p}'(\mathbf{T}_i, \mathbf{T}_j, d, \mathbf{c})]\!-\!b_j\big) - \frac{t_je^{a_j}}{t_ie^{a_i}} \bigl(I_i[\mathbf{p}]\!-\!b_i\bigr),
	\end{align}
	where $(\mathbf{T}_i, \mathbf{T}_j, d, \mathbf{c},a_i,a_j,b_i,b_j) := \boldsymbol{x} \bp \boldsymbol{\zeta}_0$ are the 
	current state variables the  residual depends on. 
	The Jacobian $\mathbf{J}_k$ is evaluated with respect to an \textit{additive increment} to $\boldsymbol{x}$, i.e.,
	\begin{align}
		\label{eq:Jac1}
		\mathbf{J}_k = 
		\frac{\partial r_k((\boldsymbol\delta + \boldsymbol{x}) \bp \boldsymbol{\zeta}_0)}{\partial \boldsymbol\delta}.
	\end{align}
	It can be decomposed as 
	\begin{align}
		\label{eq:Jac2}
		\mathbf{J}_k = 
		\biggl[ 
		\underbrace{\frac{\partial I_j}{\partial \mathbf{p}'}}_{\mathbf{J}_I}
		\underbrace{\frac{\partial \mathbf{p}'((\boldsymbol\delta\!+\!\boldsymbol{x}) \bp \boldsymbol{\zeta}_0)}{\partial \boldsymbol\delta_\text{geo}}}_{\mathbf{J}_\text{geo}}
		,~~
		\underbrace{\frac{\partial r_k((\boldsymbol\delta\!+\!\boldsymbol{x}) \bp \boldsymbol{\zeta}_0)}{\partial \boldsymbol \delta_\text{photo}}}_{\mathbf{J}_\text{photo}}
		\biggr],
	\end{align}
	where $\boldsymbol\delta_\text{geo}$ denotes the \quotes{geometric} parameters $(\mathbf{T}_i, \mathbf{T}_j, d, \mathbf{c)}$, and
	$\boldsymbol\delta_\text{photo}$ denotes the \quotes{photometric} parameters $(a_i, a_j, b_i, b_j)$.
	We employ two approximations, described below.
	
	First, both $\mathbf{J}_\text{photo}$ and $\mathbf{J}_\text{geo}$ are evaluated 
	at $\boldsymbol{x}=0$. This technique is called \quotes{First Estimate Jacobians} \cite{leutenegger2015ijrr,huang200iser}, 
	and is required to maintain consistency of the system and prevent the accumulation of spurious information. 
	In particular, in the presence of non-linear null-spaces in the energy (in our formulation 
	absolute pose and scale), adding linearizations around different evaluation points eliminates these 
	and thus slowly corrupts the system. In practice, this approximation is very good, since 
	$\mathbf{J}_\text{photo}$, $\mathbf{J}_\text{geo}$ are smooth compared to the size of the increment $\boldsymbol{x}$.
	In contrast, $\mathbf{J}_I$ is much less smooth, but does not affect the null-spaces. 
	Thus, it is evaluated at the current value for $\boldsymbol{x}$, i.e., at the same point as the residual $r_k$.
	We use centred differences to compute the image derivatives at integer positions, which are then bilinearly interpolated.
	
	Second, $\mathbf{J}_\text{geo}$ is assumed to be the same for all residuals belonging to the same point, 
	and evaluated only for the center pixel. Again, this approximation is very good in practice. While it 
	significantly reduces the required computations, we have not observed a notable effect on accuracy for 
	any of the used datasets.

	From the resulting linear system, an increment is computed as $\boldsymbol\delta = \mathbf{H}^{-1} \mathbf{b}$ and added to the current state:
	\begin{align}
		\boldsymbol{x}^\text{new} \leftarrow \boldsymbol\delta + \boldsymbol{x}. 
	\end{align}	
	Note that due to the First Estimate Jacobian approximation, a multiplicative formulation (replacing $(\boldsymbol\delta\!+\!\boldsymbol{x}) \bp \boldsymbol{\zeta}_0$ with $\boldsymbol\delta \bp (\boldsymbol{x} \bp \boldsymbol{\zeta}_0)$ in  (\ref{eq:Jac1})) results in the exact same Jacobian,
	thus a multiplicative update step $\boldsymbol{x}^\text{new} \leftarrow \log (\boldsymbol\delta \bp e^{\boldsymbol{x}} )$ is equally valid.
		
	After each update step, we update $\boldsymbol{\zeta}_0$ \textit{for all variables that are not part of the marginalization term}, using
	$\boldsymbol{\zeta}_0^\text{new}~\leftarrow~\boldsymbol{x}~\bp~\boldsymbol{\zeta}_0$ and $\boldsymbol{x}~\leftarrow 0$. In 
	practice, this includes all depth values, as well as the pose of the newest keyframe.
	Each time a new keyframe is added, we perform up to 6 Gauss-Newton iterations, breaking early if $\boldsymbol\delta$ is sufficiently small. 
	We found that -- since we never start far-away from the minimum -- a Levenberg-Marquad dampening 
	(which slows down convergence) is not required.

	\paragraph{Marginalization.} 
	\label{par:Marg}
	When the active set of variables becomes too large, old variables are removed by marginalization using the Schur complement. 
	Similar to \cite{leutenegger2015ijrr}, we drop any residual terms that would affect the sparsity pattern of $\mathbf{H}$: 
	When marginalizing frame $i$, we first marginalize all points in $\mathcal{P}_i$, as well as points that 
	have not been observed in the last two keyframes. Remaining observations of active points in 
	frame $i$ are dropped from the system. 
	
	Marginalization proceeds as follows: Let $E'$ denote the part of the energy containing all residuals that
	depend on state variables to be marginalized. We first compute a Gauss-Newton approximation of $E'$ 
	around the current state estimate $\boldsymbol{\zeta} = \boldsymbol{x} \bp \boldsymbol{\zeta}_0$. This gives
	\begin{align}
		&E'(\boldsymbol{x} \bp \boldsymbol{\zeta}_0) \\ \nonumber
		&\approx 2 (\boldsymbol{x}\!-\!\boldsymbol{x}_0)^T \mathbf{b} + (\boldsymbol{x}\!-\!\boldsymbol{x}_0)^T \mathbf{H} (\boldsymbol{x}\!-\!\boldsymbol{x}_0) + c\\ \nonumber
		&= 2 \boldsymbol{x}^T \underbrace{(\mathbf{b}\!-\!\mathbf{H} \boldsymbol{x}_0)}_{=: \mathbf{b}'} + \boldsymbol{x}^T \mathbf{H} \boldsymbol{x} + \underbrace{(c\!+\!\boldsymbol{x}_0^T \mathbf{H} \boldsymbol{x}_0\!-\!\boldsymbol{x}_0^T \mathbf{b})}_{=: c'},
	\end{align}
	where $\boldsymbol{x}_0$ denotes the current value (evaluation point for $\mathbf{r}$) of $\boldsymbol{x}$. The constants $c, c'$ can be dropped, and $\mathbf{H}, \mathbf{b}$ are defined as in (\ref{eq:GNApprox}-\ref{eq:Jac2}). This is a quadratic
	function on $\boldsymbol{x}$, and we can apply the Schur complement to marginalize a subset of variables. Written as a linear system, 
	it becomes 
	\begin{align}
		\begin{bmatrix} \mathbf{H}_{\alpha\alpha} & \mathbf{H}_{\alpha\beta}\\\mathbf{H}_{\beta\alpha} & \mathbf{H}_{\beta\beta}\end{bmatrix}
		\begin{bmatrix} {{\boldsymbol{x}}}_{\alpha} \\{{\boldsymbol{x}}}_{\beta}\end{bmatrix}
		=
		\begin{bmatrix} \mathbf{b}'_{\alpha} \\\mathbf{b}'_{\beta}\end{bmatrix},
	\end{align}
	where $\beta$ denotes the block of variables we would like to marginalize, and $\alpha$ the block of 
	variables we would like to keep. Applying the Schur complement yields \mbox{$\widehat{\mathbf{H}_{\alpha\alpha}} \boldsymbol{x}_{\alpha} = \widehat{\mathbf{b}'_{\alpha}}$}, with
	\begin{align}
		\widehat{\mathbf{H}_{\alpha\alpha}} &= \mathbf{H}_{\alpha\alpha} - \mathbf{H}_{\alpha\beta} \mathbf{H}_{\beta\beta}^{-1} \mathbf{H}_{\beta\alpha}\\
		\widehat{\mathbf{b}'_{\alpha}} &= \mathbf{b}'_{\alpha} - \mathbf{H}_{\alpha\beta} \mathbf{H}_{\beta\beta}^{-1} \mathbf{b}'_{\beta}.
	\end{align}
	The residual energy on $\boldsymbol{x}_{\alpha}$ can hence be written as
	\begin{align}
		E'\!\big(\boldsymbol{x}_{\alpha} \bp (\!{\boldsymbol{\zeta}_0}\!)_{\alpha}\big) = 2 \boldsymbol{x}_{\alpha}^T \widehat{\mathbf{b}'_{\alpha}} + \boldsymbol{x}_{\alpha}^T \widehat{\mathbf{H}_{\alpha\alpha}} \boldsymbol{x}_{\alpha}.
	\end{align}
	This is a quadratic function on $\boldsymbol{x}$ and can be trivially added to the full photometric error $E_\text{photo}$ during all subsequent optimization and 
	marginalization operations, replacing the corresponding non-linear terms. Note that this requires the tangent space 
	for $\boldsymbol{\zeta}_0$ to remain the same for all variables that appear in $E'$ during all subsequent 
	optimization and marginalization steps.

	\section{Visual Odometry Front-End}
	\label{secFrontEnd}
	The front end is the part of the algorithm that
	\begin{itemize}
	\item determines the sets $\mathcal{F}, \mathcal{P}_i$, and $\text{obs}(\mathbf{p})$ that make up the error 
	terms of $E_\text{photo}$. It decides which points and frames are used,
	and in which frames a point is visible -- in particular, this includes outlier removal
	and occlusion detection. 
	\item provides initializations for new parameters, required for optimizing the
	highly non-convex energy function $E_\text{photo}$. As a rule of thumb, a linearization of the image 
	$I$ is only valid in a 1-2 pixel radius; hence all parameters involved in computing $\mathbf{p}'$ 
	should be initialized sufficiently accurately for $\mathbf{p}'$ to be off by no more than 1-2 pixels.
	\item decides when a point / frame should be marginalized.
	\end{itemize}
	As such, the front-end needs to replace many operations that in the indirect setting are accomplished by
	keyframe detectors (determining visibility, point selection) and initialization procedures such as RANSAC. Note that many
	procedures described here are specific to the monocular case. For instance, using a stereo camera
	makes obtaining initial depth values more straightforward, while integration of an IMU can significantly
	robustify -- or even directly provide -- a pose initialization for new frames.

	\subsection{Frame Management}
	\label{sssecFrameManagement}
	Our method always keeps a window of up to $N_f$ active keyframes (we use $N_f=7$). Every new
	frame is initially tracked with respect to these reference frames (Step 1). It is then either discarded
	or used to create a new keyframe (Step 2). Once a new keyframe -- and respective new points -- are 
	created, the total photometric error (\ref{eq:FullPhotErr}) is optimized. Afterwards,
	we marginalize one or more frames (Step 3). 
	
	\begin{figure}
\centering
\includegraphics[width=.245\linewidth]{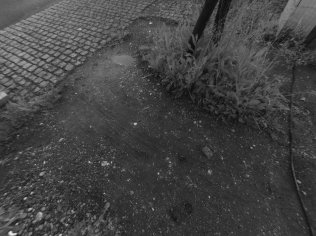}\hspace{-0.5mm}
\includegraphics[width=.245\linewidth]{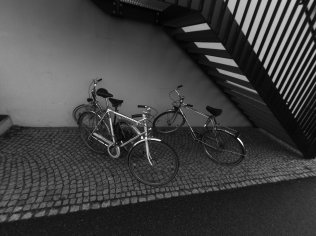}\hspace{-0.5mm}
\includegraphics[width=.245\linewidth]{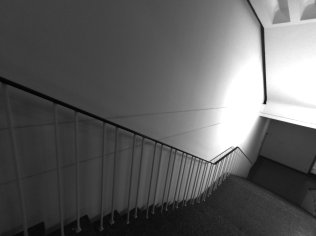}\hspace{-0.5mm}
\includegraphics[width=.245\linewidth]{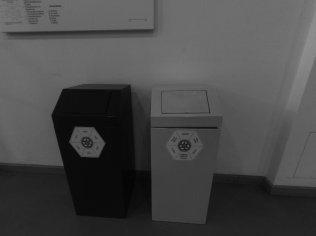}\\
\includegraphics[width=.245\linewidth]{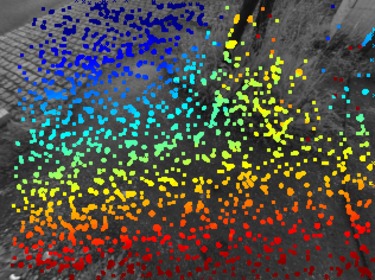}\hspace{-0.5mm}
\includegraphics[width=.245\linewidth]{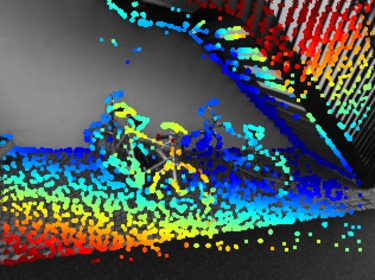}\hspace{-0.5mm}
\includegraphics[width=.245\linewidth]{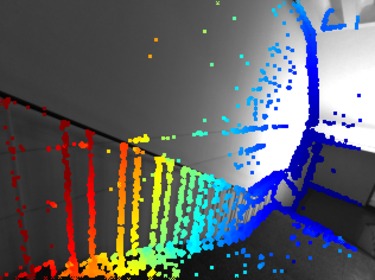}\hspace{-0.5mm}
\includegraphics[width=.245\linewidth]{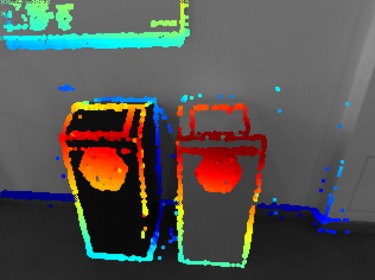}\\
\caption{\textbf{Example depth maps used for initial frame tracking.} The top row shows the original images, the bottom row the color-coded depth maps. Since we aim at a fixed number of points in the active optimization, 
they become more sparse in densely textured scenes (left), while becoming similar in density to those of LSD-SLAM in scenes where 
only few informative image regions are available to sample from (right).}
\label{fig:sparseMapsExample}
\end{figure}

	\paragraph{Step 1: Initial Frame Tracking.} When a new keyframe is created, all active points are projected into it
	and slightly dilated, creating a semi-dense depth map. New frames are tracked with respect 
	to only this frame using conventional two-frame direct image alignment, a multi-scale image pyramid and a constant motion model
	to initialize. 
	Figure~\ref{fig:sparseMapsExample} shows some examples -- we found that further increasing the density has little to no benefit in terms of 
	accuracy or robustness, while significantly increasing runtime. Note that when down-scaling the images, a pixel is 
	assigned a depth value if at least one of the source pixels has a depth value as in \cite{schoeps14ismar}, 
	significantly increasing the density on coarser resolutions.
	
	If the final RMSE for a frame is more than twice that of the frame before, we assume that direct 
	image alignment failed and attempt to recover by initializing with up to 27 different small rotations in different directions. 
	This recovery-tracking is done on the coarsest pyramid level only, and takes approximately 0.5ms per try.
	Note that this RANSAC-like procedure is only rarely invoked, such as when the camera moves very quickly shakily. Tightly
	integrating an IMU would likely render this unnecessary.

	\paragraph{Step 2: Keyframe Creation.}
	Similar to ORB-SLAM, our strategy is to initially take many keyframes (around 5-10 keyframes per second), 
	and sparsify them afterwards by early marginalizing redundant keyframes. We combine three criteria to determine if
	a new keyframe is required:
	\begin{enumerate} 
		\item New keyframes need to be created as the field of view changes. We measure this
		by the mean square optical flow (from the last keyframe to the latest frame)
		\mbox{$f := (\frac{1}{n} \sum_{i=1}^n \|\mathbf{p}-\mathbf{p}'\|^2)^\frac{1}{2}$} during initial coarse tracking.
		\item Camera translation causes occlusions and dis-occlusions,
		which requires more keyframes to be taken (even though $f$ may be small).
		This is measured by the mean flow without rotation, i.e., 
		\mbox{$f_t := (\frac{1}{n} \sum_{i=1}^n \|\mathbf{p}-\mathbf{p}_t'\|^2)^\frac{1}{2}$}, where $\mathbf{p}_t$ is the warped point position with $\mathbf{R}=\mathbf{I}_{3\times 3}$.
		\item If the camera exposure time changes significantly, a new keyframe should be 
		taken. This is measured by the relative brightness factor between two frames \mbox{$a := |\log(e^{a_j-a_i} t_j t_i^{-1})|$}.
	\end{enumerate}
	These three quantities can be obtained easily as a by-product of initial alignment. Finally, a new keyframe is taken if 
	\mbox{$w_f f + w_{f_t} f_t + w_a a > T_\text{kf}$}, where $w_f, w_{f_t}, w_a$ provide a relative weighting of these three 
	indicators, and $T_\text{kf}=1$ by default.

\begin{figure}
\centering
{\setlength{\fboxsep}{0pt}\setlength{\fboxrule}{0.5pt}\fbox{\includegraphics[width=.982\linewidth]{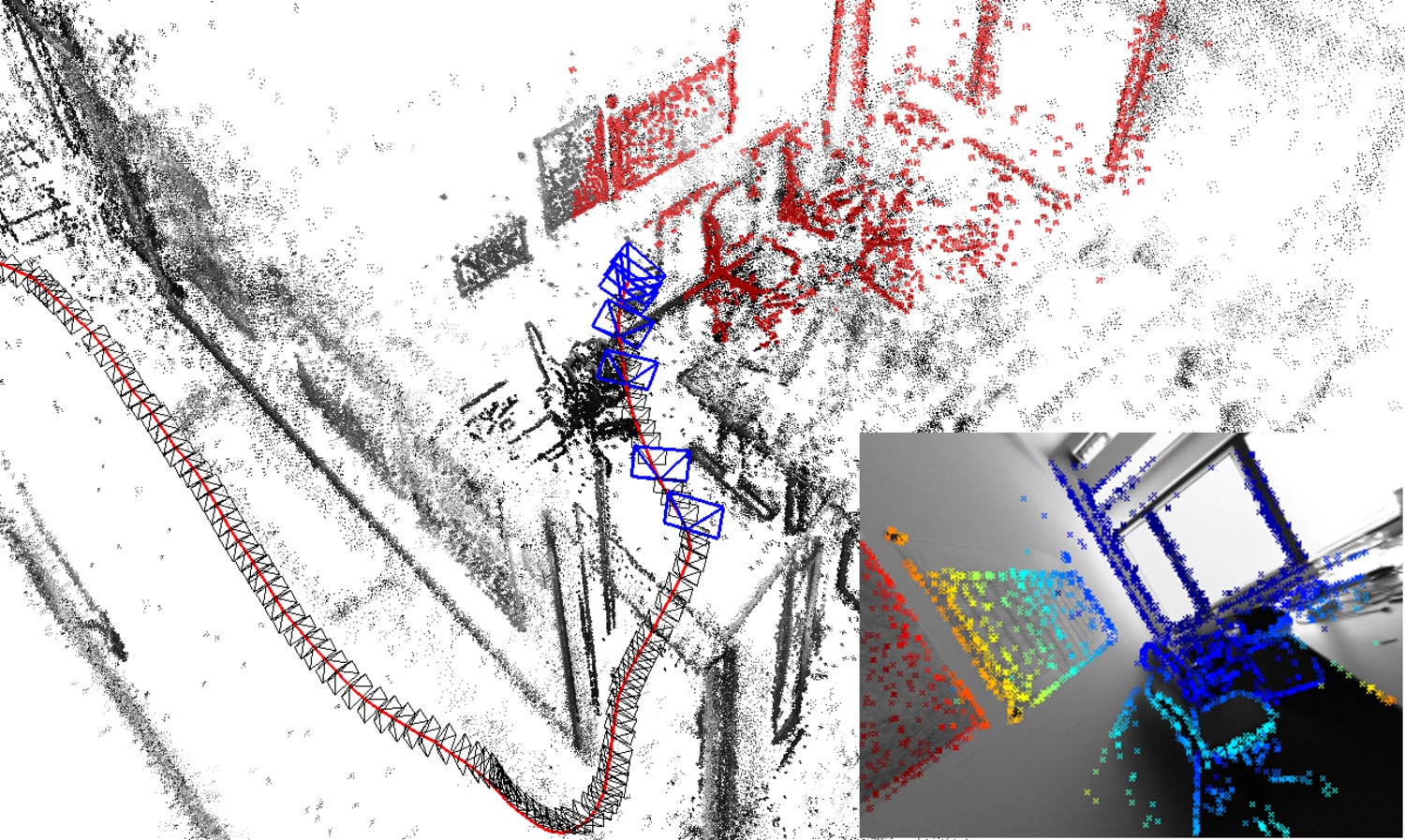}}}\\
\includegraphics[width=.325\linewidth]{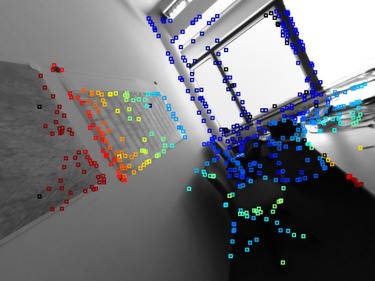}\hspace{-.5mm}
\includegraphics[width=.325\linewidth]{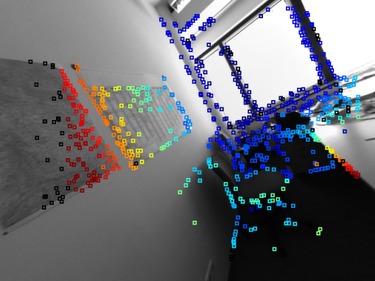}\hspace{-.5mm}
\includegraphics[width=.325\linewidth]{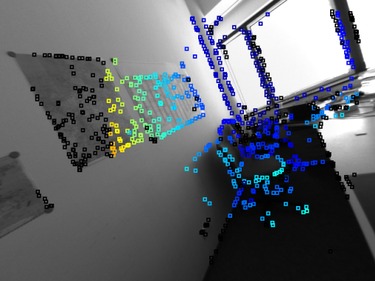}\\
\includegraphics[width=.325\linewidth]{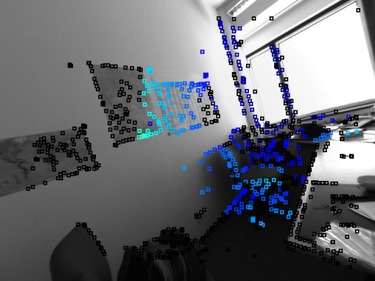}\hspace{-.5mm}
\includegraphics[width=.325\linewidth]{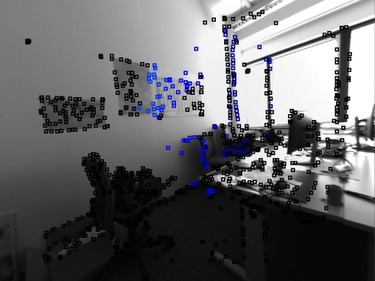}\hspace{-.5mm}
\includegraphics[width=.325\linewidth]{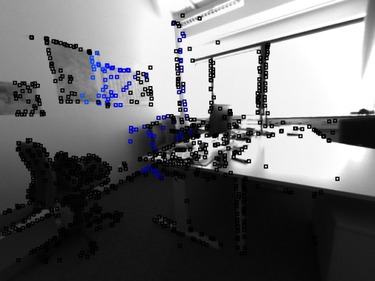}\\
\caption{\textbf{Keyframe management.} Bottom rows: The 6 old keyframes in the optimization window, overlaid
with the points hosted in them (already marginalized points are shown in black).
The top image shows the full point cloud, as well as the positions of
all keyframes (black camera frustums) -- active points and keyframes are shown in red and blue respectively. 
The inlay shows the newly added keyframe, overlaid with all forward-warped active points, which will be used
for initial alignment of subsequent frames.}
\label{fig:KFMarg}
\end{figure}

	\paragraph{Step 3: Keyframe Marginalization.} Our marginalization strategy is as follows (let $I_1 \hdots I_n$ be the set of active keyframes, with $I_1$ being the newest and $I_n$ being the oldest):
	\begin{enumerate}
		\item We always keep the latest two keyframes ($I_1$ and $I_2$). 
		\item Frames with less than 5\% of their points visible in $I_1$ are marginalized.
		\item If more than $N_f$ frames are active, we marginalize the one (excluding $I_1$ and $I_2$) which 
		maximizes a \quotes{distance score} $s(I_i)$, computed as
		\begin{align}
			s(I_i) = \sqrt{d(i, 1)} \sum_{j \in [3, n] \backslash \{i\}} (d(i, j) + \epsilon)^{-1},
		\end{align}
		where $d(i, j)$ is the Euclidean distance between keyframes $I_i$ and $I_j$, and $\epsilon$ a 
		small constant. This scoring function is heuristically designed
		to keep active keyframes well-distributed in 3D space, with more keyframes close to the most recent one.
	\end{enumerate}
	A keyframe is marginalized by first marginalizing all points represented in it, and then the frame itself, using the marginalization procedure from Section~\ref{par:Marg}.
	To preserve the sparsity structure of the Hessian, all observations of still existing points in the frame are dropped from the system.
	 While this is clearly suboptimal (in practice about half of 
	all residuals are dropped for this reason), it allows to efficiently optimize the energy function.
	Figure~\ref{fig:KFMarg} shows an example of a scene, highlighting the active set of points and frames.
	
	\subsection{Point Management} 
	\label{sssecpointManagement}
	Most existing direct methods focus on utilizing as much image data as possible. To achieve this in real time,
	they accumulate early, sub-optimal estimates (linearizations / depth triangulations), and ignore -- or approximate -- correlations between different parameters. 
	In this work, we follow a different approach, and instead heavily sub-sample data to allow processing it in real time 
	in a joint optimization framework. In fact, our experiments show that image data is highly redundant, and the benefit of
	simply using \textit{more} data points quickly flattens off. 
	Note that in contrast to indirect methods, our direct framework still allows to \textit{sample from across all available data},
	including weakly textured or repetitive regions and edges, which does provide a real benefit (see Section~\ref{secResults}).
	
	We aim at always keeping a fixed number $N_p$ of active points (we use $N_p=2000$), equally distributed across space 
	and active frames, in the optimization.
	In a first step, we identify $N_p$ candidate points in each new keyframe (Step 1). 
	Candidate points are not immediately added into the optimization, but instead are tracked 
	individually in subsequent frames, generating a coarse depth value which will serve as initialization (Step 2). 
	When new points need to be added to the optimization, we choose a number of candidate points (from across
	all frames in the optimization window) to be activated, i.e., added into the optimization (Step 3).
	Note that we choose $N_p$ candidates \textit{in each frame}, however only keep $N_p$ active 
	points \textit{across all active frames combined}. This assures that we always have sufficient candidates 
	to activate, even though some may become invalid as they leave the field of view or are identified as outliers.

	\paragraph{Step 1: Candidate Point Selection.} Our point selection strategy aims at selecting points that are (1) well-distributed in the image 
	and (2) have sufficiently high image gradient magnitude with respect to their immediate surroundings. 
	We obtain a region-adaptive gradient threshold by splitting the image into $32\times32$ blocks. For each block, we then compute
	the threshold as $\bar{g} + g_\text{th}$, where $\bar{g}$ is the median absolute gradient over all pixels in that block, and $g_\text{th}$ 
	a global constant (we use $g_\text{th}=7$).
	
\begin{figure}
\centering
\includegraphics[width=.245\linewidth]{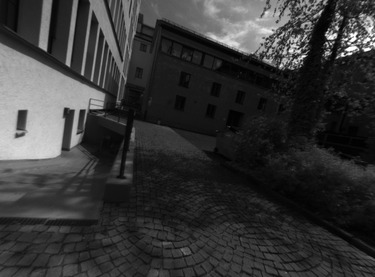}\hspace{-0.5mm}
\includegraphics[width=.245\linewidth]{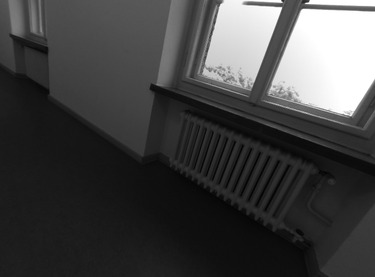}\hspace{-0.5mm}
\includegraphics[width=.245\linewidth]{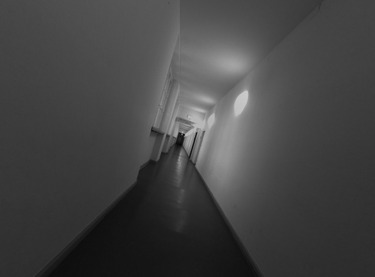}\hspace{-0.5mm}
\includegraphics[width=.245\linewidth]{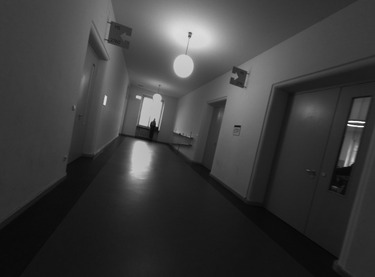}\\
\includegraphics[width=.245\linewidth]{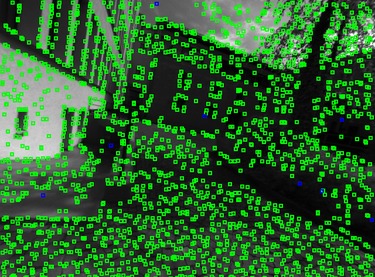}\hspace{-0.5mm}
\includegraphics[width=.245\linewidth]{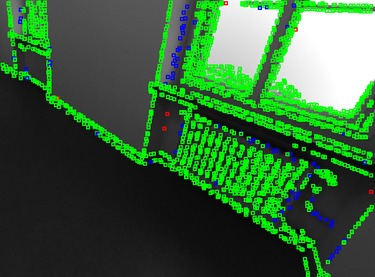}\hspace{-0.5mm}
\includegraphics[width=.245\linewidth]{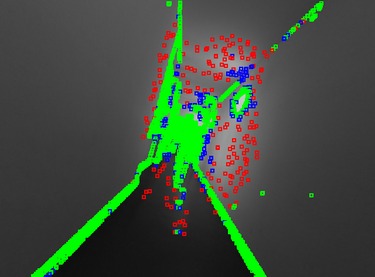}\hspace{-0.5mm}
\includegraphics[width=.245\linewidth]{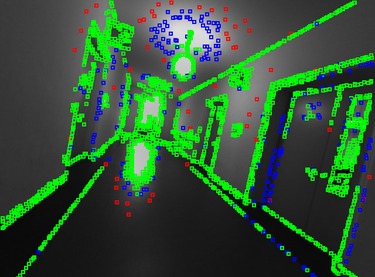}\\
\caption{\textbf{Candidate selection.} 
The top row shows the original images, the bottom row shows the points chosen as candidates to be added to the map (2000 in each frame). 
Points selected on the first pass are shown in green, those selected on the second and third pass in blue and red respectively. 
Green candidates are evenly spread across gradient-rich areas, while points added on the second and third 
pass also cover regions with very weak intensity variations, but are much sparser.}
\label{fig:pointSelectionExample}
\end{figure}

	To obtain an equal distribution of points throughout the image, we split it into $d \times d$ blocks, and from each block select the pixel with largest gradient if it surpasses the region-adaptive threshold. 
	Otherwise, we do not select a pixel from that block. 
	We found that it is often beneficial to also include some points with weaker gradient from regions where no high-gradient points are present,
	capturing information from weak intensity variations originating for example from smoothly changing illumination across white walls.
	To achieve this, we repeat this procedure twice more, with decreased gradient threshold and block-size $2d$ and $4d$, respectively.
	 The block-size $d$ is continuously adapted such that this procedure generates the desired amount of points (if too many points were created 
	it is increased for the next frame, otherwise it is decreased). Figure~\ref{fig:pointSelectionExample} shows the selected point candidates for some example scenes.
	Note that for for candidate point selection, we use the raw images prior to photometric correction.

	\paragraph{Step 2: Candidate Point Tracking.} Point candidates are tracked in subsequent frames using a discrete search along the epipolar 
	line, minimizing the photometric error (\ref{eq:PixelPhotErr}). From the best match we compute a depth and associated variance,
	which is used to constrain the search interval for the subsequent frame. 
	This tracking strategy is inspired by LSD-SLAM. Note that the computed depth only serves as \textit{initialization} once
	the point is activated.

	\paragraph{Step 3: Candidate Point Activation.} 
	After a set of old points is marginalized, new point candidates are activated to replace them. Again, we aim at maintaining a uniform 
	spatial distribution across the image. To this end, we first project all active points onto the most recent keyframe. We then 
	activate candidate points which -- also projected into this keyframe -- maximize the distance to any existing point (requiring larger distance for candidates created during the second or third block-run). Figure~\ref{fig:sparseMapsExample} shows the resulting distribution of points in a number of scenes.

	\paragraph{Outlier and Occlusion Detection.}
	Since the available image data generally contains much more information than can be used in real time, 
	we attempt to identify and remove potential outliers as early as possible. 
	First, when searching along the epipolar line during candidate tracking, points for which 
	the minimum is not sufficiently distinct are permanently discarded, greatly reducing the number of false matches in 
	repetitive areas. 
	Second, point observations for which the photometric error (\ref{eq:PixelPhotErr}) surpasses a threshold 
	are removed. The threshold is continuously adapted with respect to the median residual in the respective 
	frame. For \quotes{bad} frames (e.g., frames that contain a lot of motion blur), the threshold will be higher, 
	such that not all observations are removed. 
	For good frames, in turn, the threshold will be lower, as we can afford to be more strict.

	\section{Results}
	\label{secResults}
\begin{figure*}
\centering
\imgOverlay{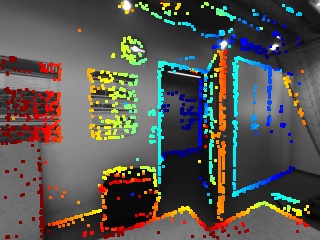}\hspace{-.5mm}%
\imgOverlay{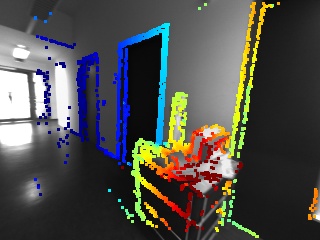}\hspace{-.5mm}%
\imgOverlay{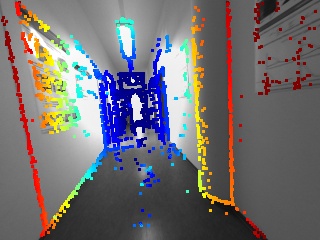}\hspace{-.5mm}%
\imgOverlay{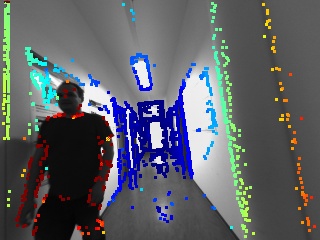}\hspace{-.5mm}%
\imgOverlay{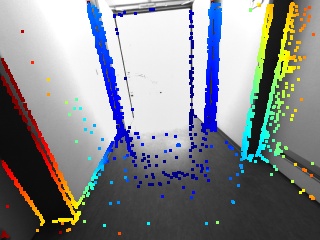}\hspace{-.5mm}%
\imgOverlay{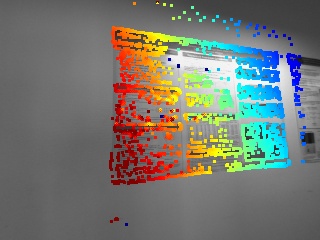}\hspace{-.5mm}%
\imgOverlay{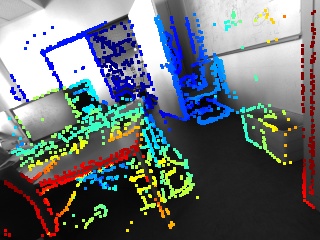}\hspace{-.5mm}%
\imgOverlay{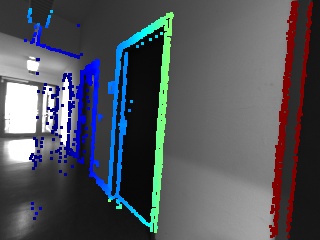}\hspace{-.5mm}%
\imgOverlay{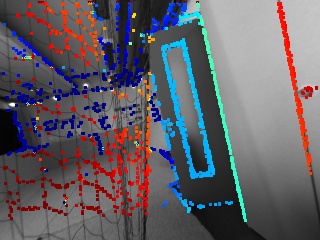}\hspace{-.5mm}%
\imgOverlay{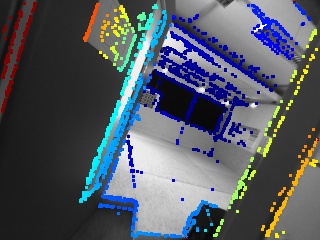}\\[.4mm]
\imgOverlay{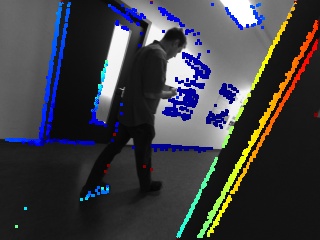}\hspace{-.5mm}%
\imgOverlay{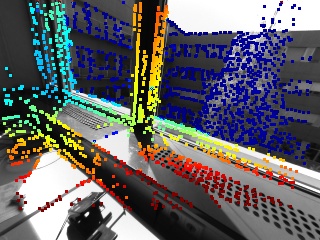}\hspace{-.5mm}%
\imgOverlay{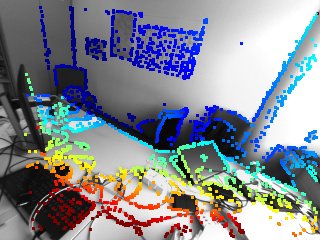}\hspace{-.5mm}%
\imgOverlay{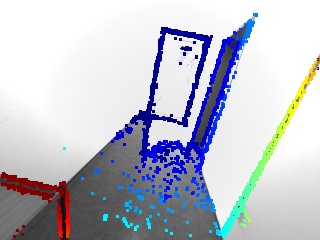}\hspace{-.5mm}%
\imgOverlay{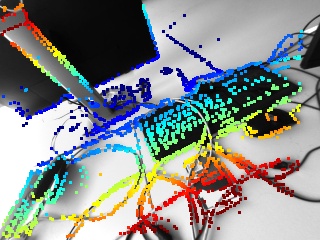}\hspace{-.5mm}%
\imgOverlay{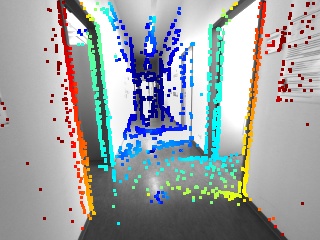}\hspace{-.5mm}%
\imgOverlay{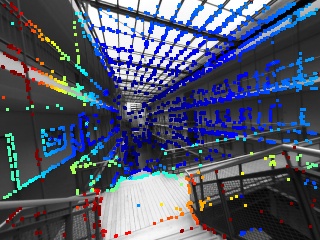}\hspace{-.5mm}%
\imgOverlay{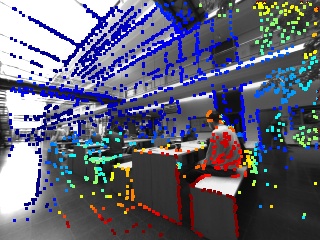}\hspace{-.5mm}%
\imgOverlay{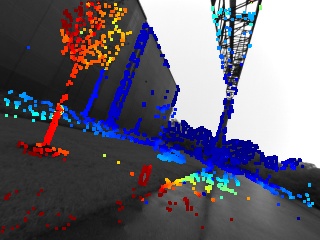}\hspace{-.5mm}%
\imgOverlay{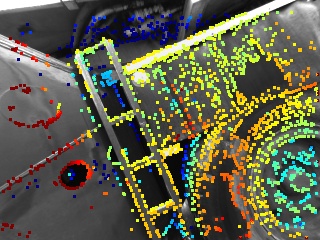}\\[.4mm]
\imgOverlay{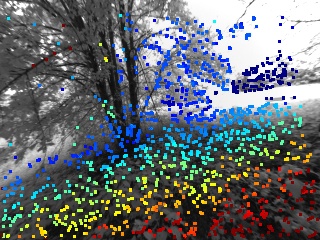}\hspace{-.5mm}%
\imgOverlay{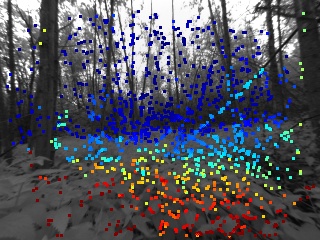}\hspace{-.5mm}%
\imgOverlay{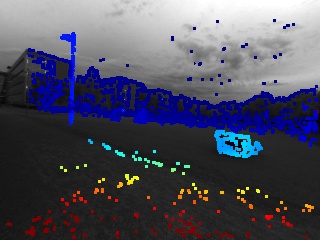}\hspace{-.5mm}%
\imgOverlay{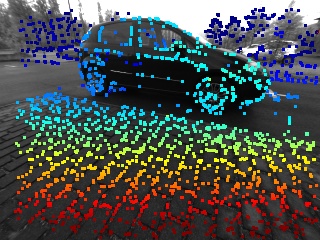}\hspace{-.5mm}%
\imgOverlay{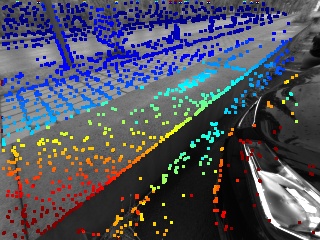}\hspace{-.5mm}%
\imgOverlay{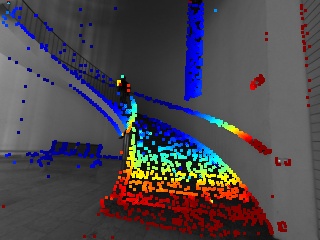}\hspace{-.5mm}%
\imgOverlay{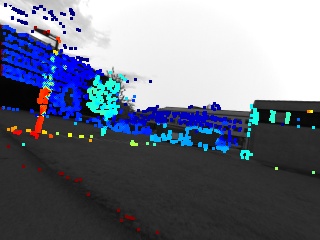}\hspace{-.5mm}%
\imgOverlay{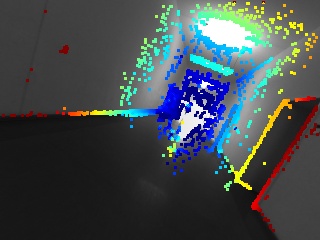}\hspace{-.5mm}%
\imgOverlay{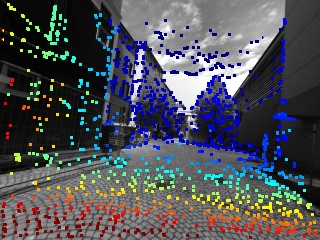}\hspace{-.5mm}%
\imgOverlay{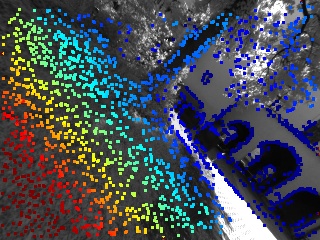}\\[.4mm]
\imgOverlay{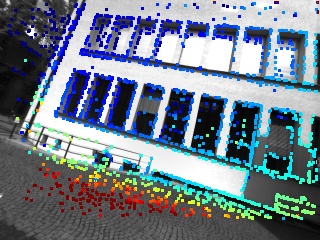}\hspace{-.5mm}%
\imgOverlay{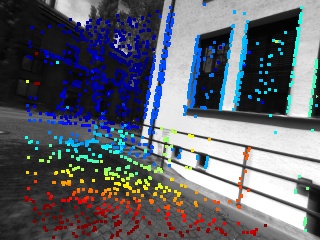}\hspace{-.5mm}%
\imgOverlay{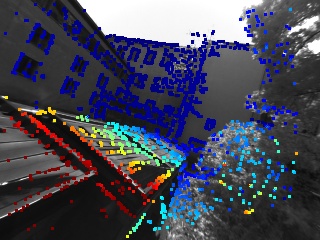}\hspace{-.5mm}%
\imgOverlay{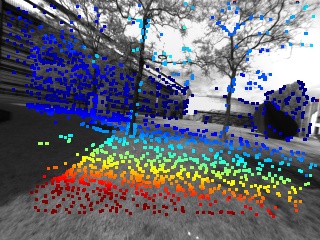}\hspace{-.5mm}%
\imgOverlay{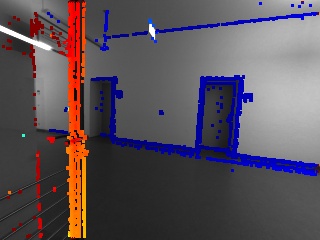}\hspace{-.5mm}%
\imgOverlay{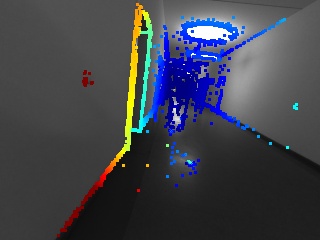}\hspace{-.5mm}%
\imgOverlay{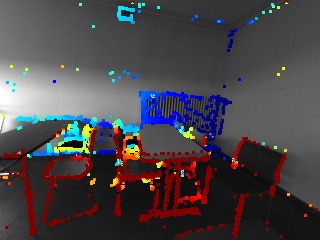}\hspace{-.5mm}%
\imgOverlay{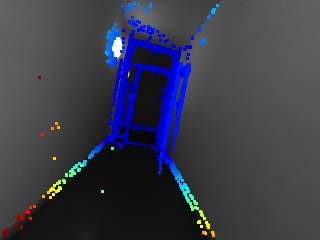}\hspace{-.5mm}%
\imgOverlay{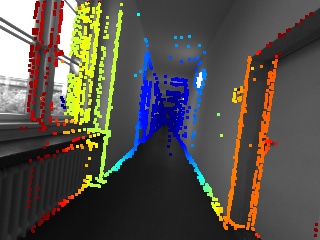}\hspace{-.5mm}%
\imgOverlay{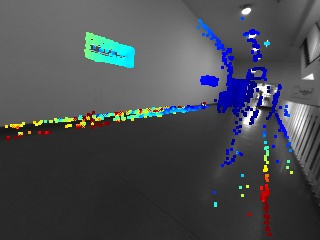}\\[.4mm]
\imgOverlay{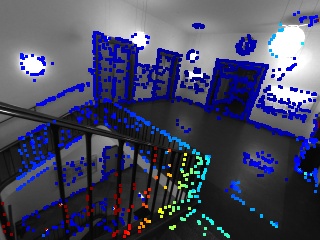}\hspace{-.5mm}%
\imgOverlay{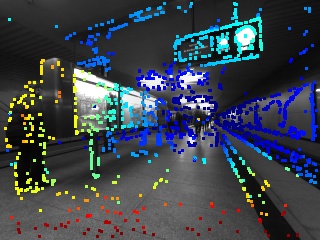}\hspace{-.5mm}%
\imgOverlay{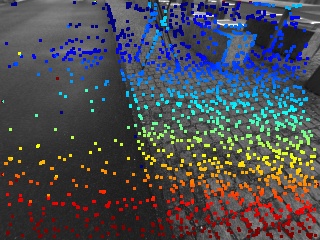}\hspace{-.5mm}%
\imgOverlay{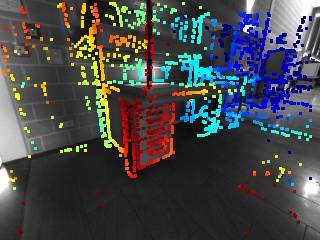}\hspace{-.5mm}%
\imgOverlay{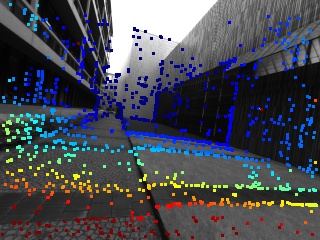}\hspace{-.5mm}%
\imgOverlay{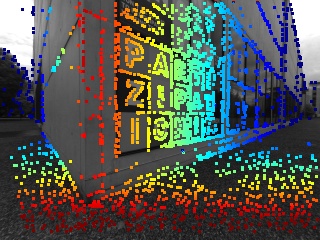}\hspace{-.5mm}%
\imgOverlay{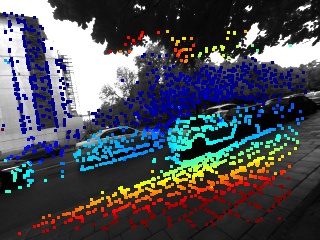}\hspace{-.5mm}%
\imgOverlay{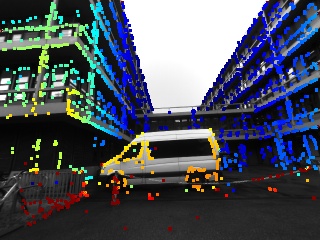}\hspace{-.5mm}%
\imgOverlay{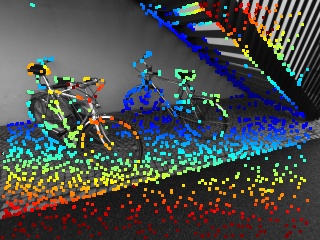}\hspace{-.5mm}%
\imgOverlay{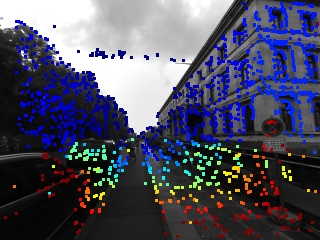}\\
\caption{\textbf{TUM mono-VO Dataset.} 
\hl{A single image from each of the 50 TUM mono-VO dataset sequences (s\_01 to s\_50) used for evaluation and parameter studies, overlayed with the predicted depth map from DSO. The full 
dataset contains over 105 minutes of video (190'000 frames). 
Note the wide range of environments covered, ranging from narrow indoor corridores to wide outdoor areas, including forests.}}
\label{fig:monoTUMdataset}
\end{figure*}

	In this section we will extensively evaluate our \textbf{D}irect \textbf{S}parse mono-V\textbf{O} algorithm (DSO). 
	We both compare it to other monocular SLAM / VO methods, as well as evaluate the effect of 
	important design and parameter choices.
	We use three datasets for evaluation: 
	
	(1) The \textbf{TUM monoVO dataset} \cite{engel16archiveDataset}, 
	which provides 50 photometrically calibrated sequences, comprising 105 minutes of video recorded in dozens of different environments,
	indoors and outdoors \hl{(see Figure \ref{fig:monoTUMdataset}).} Since the dataset only provides loop-closure-ground-truth (allowing to evaluate tracking accuracy via the 
	accumulated drift after a large loop), 
	we evaluate using the \textit{alignment error} ($e_\text{align}$) as defined in the respective publication.
	
	(2) The \textbf{EuRoC MAV dataset} \cite{burrii6ijrr}, which contains 11 stereo-inertial sequences comprising 19 minutes of video, 
	recorded in 3 different indoor environments. For this dataset, no photometric calibration or 
	exposure times are available, hence we omit photometric image correction and set ($\lambda_a=\lambda_b=0$).
	We evaluate in terms of the absolute trajectory error ($e_\text{ate}$), which is the translational RMSE
	after $\text{Sim}(3)$ alignment. For this dataset we crop the beginning of each sequence since they contain very
	shaky motion meant to initialize the IMU biases -- we only use the parts of the sequence where the MAV is in the air.
	
	(3) The \textbf{ICL-NUIM dataset} \cite{handa14icra}, which contains 8 ray-traced 
	sequences comprising 4.5 minutes of video, from two indoor environments. For this dataset, photometric 
	image correction is not required, and all exposure times can be set to $t=1$. 
	Again, we evaluate in terms of the absolute trajectory error ($e_\text{ate}$).

\begin{figure}
\centering
\begin{minipage}{0.03\linewidth}\rotatebox{90}{\footnotesize ~~~number of runs}\\\end{minipage}
\begin{minipage}{.91\linewidth}\centering\includegraphics[width=1\linewidth]{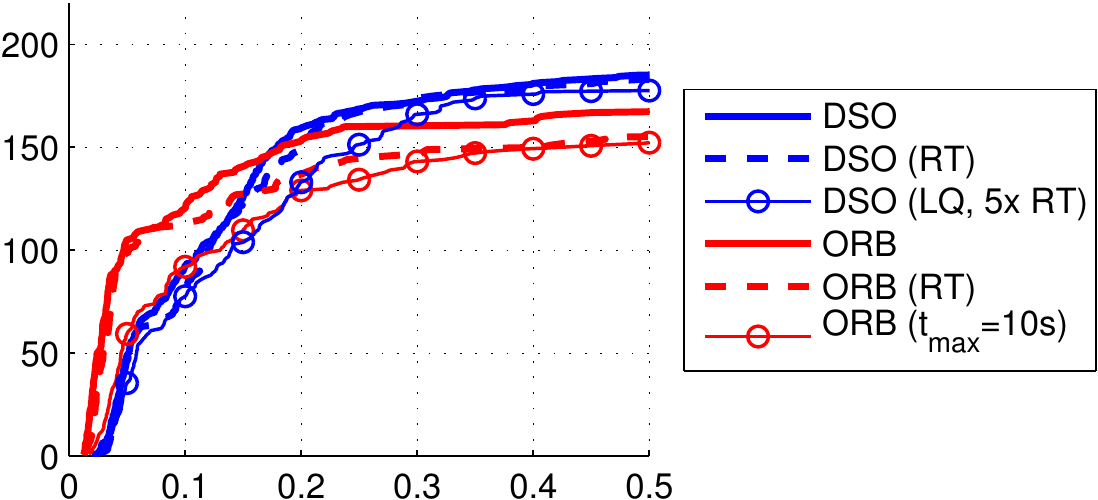}\\[-.5mm]\footnotesize  $e_\text{rmse}$\end{minipage}\\[2mm]
\begin{minipage}{0.03\linewidth}\rotatebox{90}{\footnotesize ~~~number of runs}\\\end{minipage}
\begin{minipage}{.91\linewidth}\centering\includegraphics[width=1\linewidth]{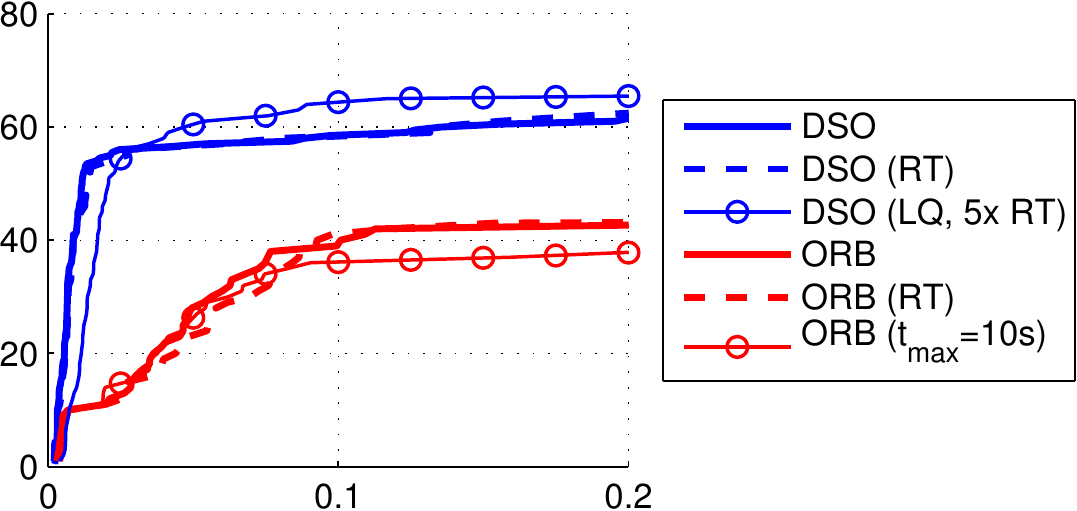}\\[-.5mm]\footnotesize  $e_\text{rmse}$\end{minipage}\\[2mm]
\caption{\textbf{Results on EuRoC MAV (top) and ICL\_NUIM (bottom) datasets.}
Translational RMSE after Sim(3) alignment. 
%We run each of the 11 sequences (a) forwards \& backwards, (b) using the left \& the right camera, and (c) five times,
%resulting in $2\cdot2\cdot 5 \cdot 11 = 220$ runs. 
RT (dashed) denotes hard-enforced real-time execution. 
Further, we evaluate DSO with low settings at 5 times real-time speed, and ORB-SLAM when restricting local loop-closures to points that have been observed at least once within the last $t_\text{max}$=10s.}
\label{fig:MAVSIM}
\end{figure}

	\paragraph{Methodology.} We aim at an evaluation as comprehensive as possible given the available data, and thus 
	run all sequences both forwards and backwards, 5 times each (to account for non-deterministic behaviour). 
	On default settings, we run each method 10 times each.
	For the EuRoC MAV dataset we further run both the left and the right video separately.
	In total, this gives 500 runs for the TUM-monoVO dataset, 220 runs for the EuRoC MAV dataset, and 80 runs for 
	the ICL-NUIM dataset, which we run on 20 dedicated workstations. We remove the dependency on the host machine's CPU speed
	by \textit{not enforcing real-time} execution, except where stated otherwise: for ORB-SLAM we play the video at 20\% speed, 
	whereas DSO is run in a sequentialized, single-threaded implementation that runs approximately four times slower than real time. 
	Note that even though we do not enforce real-time execution for most of the experiments, 
	we use the exact same parameter settings as for the real-time comparisons.
	
	The results are summarized in the form of \textit{cumulative error plots} (see, e.g., Figure~\ref{fig:MAVSIM}), which visualize 
	for how many tracked sequences the respective error value ($e_\text{ate}$ / $e_\text{align}$) was below a certain threshold; thereby showing both 
	accuracy on sequences where a method works well, as well as robustness, i.e., on how many sequences the method does not fail.
	The raw tracking results for all runs -- as well as scripts to compute the figures -- are provided in the supplementary material\footnote{\url{http://vision.in.tum.de/dso}}.
	Additional interesting analysis using the TUM-monoVO dataset -- e.g. the influence of the camera's field of view, 
	the image resolution or the camera's motion direction -- can be found in \cite{engel16archiveDataset}.

\begin{figure}
\centering
\begin{minipage}{0.03\linewidth}\rotatebox{90}{\footnotesize ~~~number of runs}\\\end{minipage}
\begin{minipage}{.45\linewidth}\centering\includegraphics[width=1\linewidth]{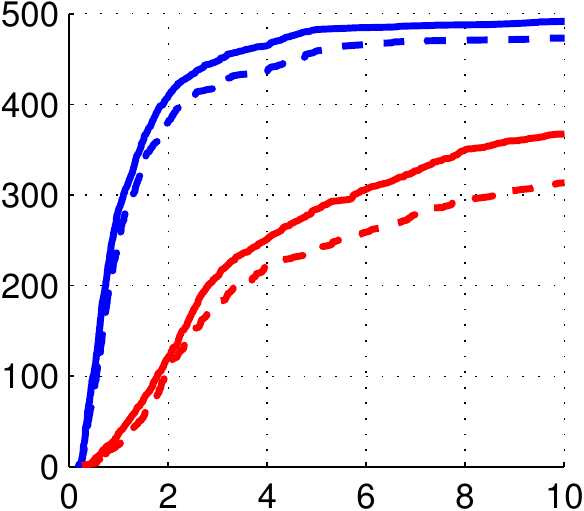}\\[-.5mm]\footnotesize ~~$e_\text{align}$\end{minipage}
\begin{minipage}{.45\linewidth}\centering\includegraphics[width=1\linewidth]{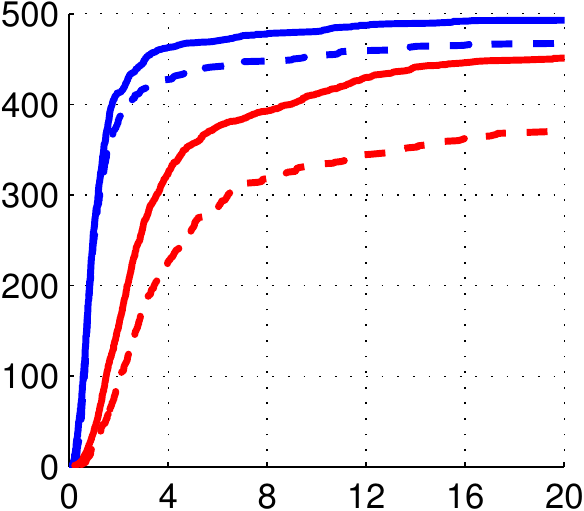}\\[-.5mm]\footnotesize ~~$e_r$ (degree)\end{minipage}\\
\begin{minipage}{0.03\linewidth}\rotatebox{90}{\footnotesize ~~~number of runs}\\\end{minipage}
\begin{minipage}{.91\linewidth}\includegraphics[width=0.95\linewidth]{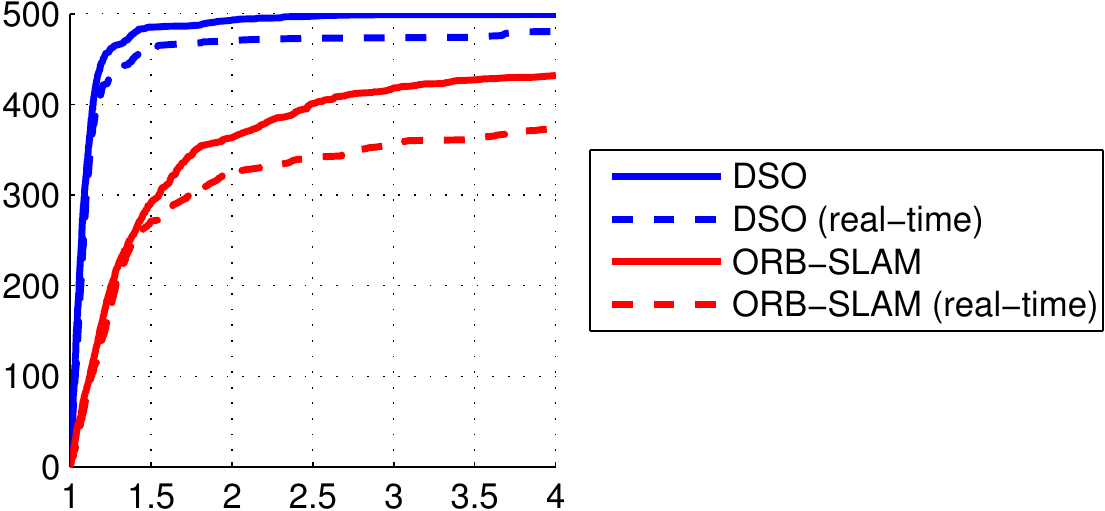}\\[-.5mm]\footnotesize {\color{white}\tiny secret Message} $e'_s$ (multiplier)\end{minipage}\\[2mm]
\caption{\textbf{Results on TUM-monoVO dataset.}
Accumulated rotational drift $e_r$ and scale drift $e_s$ after a large loop, as well as the alignment error as defined in \cite{engel16archiveDataset}. Since $e_s$ is a multiplicative factor, we aggregate $e'_s = \max(e_s, e_s^{-1})$. The solid line
corresponds to sequentialized, non-real-time execution, the dashed line to hard enforced real-time processing. For DSO, we also 
show results obtained at low parameter settings, running at 5 times real-time speed.}
\label{fig:accTUM}
\end{figure}

	\paragraph{Evaluated Methods and Parameter Settings.}
	We compare our method to the open-source implementation of (monocular) ORB-SLAM \cite{mur2015orb}. We also attempted
	to evaluate against the open-source implementations of LSD-SLAM \cite{engel14eccv} and SVO \cite{forster14icra}, 
	however both methods consistently fail on most of the sequences. A major reason for this is that 
	they assume brightness constancy (ignoring exposure changes), while both real-world datasets used contain heavy
	exposure variations.

	To facilitate a fair comparison and allow application of the loop-closure metric from the TUM-monoVO dataset,
	we disable explicit loop-closure detection and re-localization for ORB-SLAM.
	Note that everything else (including local and global BA) remains unchanged, still allowing ORB-SLAM to 
	detect incremental loop-closures that can be found via the co-visibility representation alone.
	All parameters are set to the same value across all sequences and datasets. 
	The only exception is the ICL-NUIM dataset: For this dataset we set $g_\text{th}=3$ for DSO, 
	and lower the FAST threshold for ORB-SLAM to 2, which we found to give best results.

\begin{figure}
\centering
ORB-SLAM:\\[0.5mm]
\includegraphics[height=2.6cm]{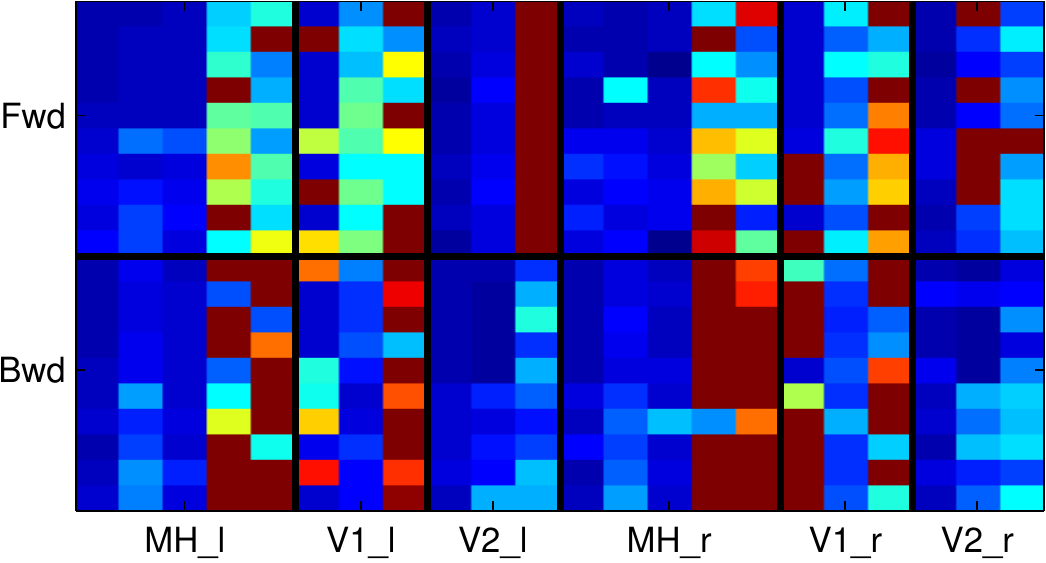}
\includegraphics[height=2.6cm]{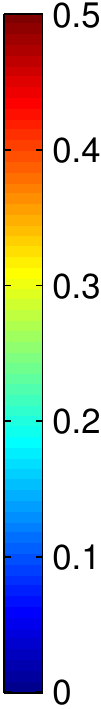}
\includegraphics[height=2.6cm]{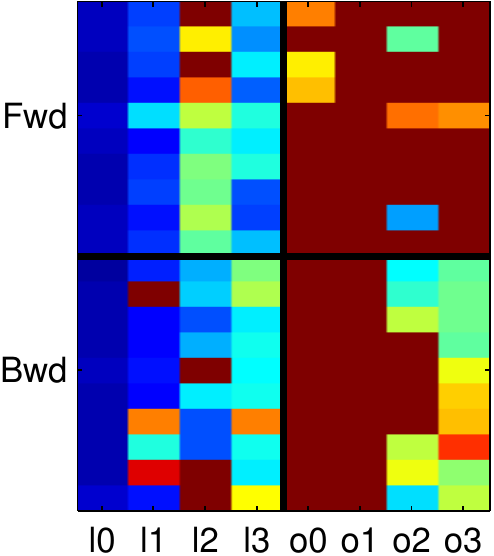}
\includegraphics[height=2.6cm]{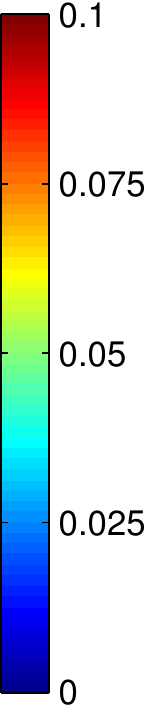}\\[1mm]
DSO:\\[0.5mm]
\includegraphics[height=2.6cm]{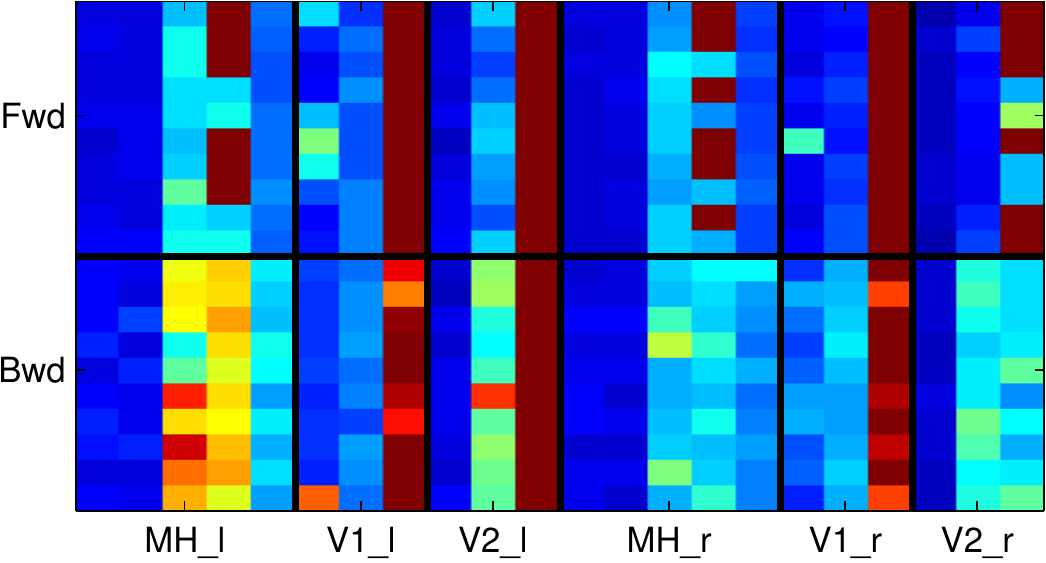}
\includegraphics[height=2.6cm]{images/results/MAV_imageLegend_crop.pdf}
\includegraphics[height=2.6cm]{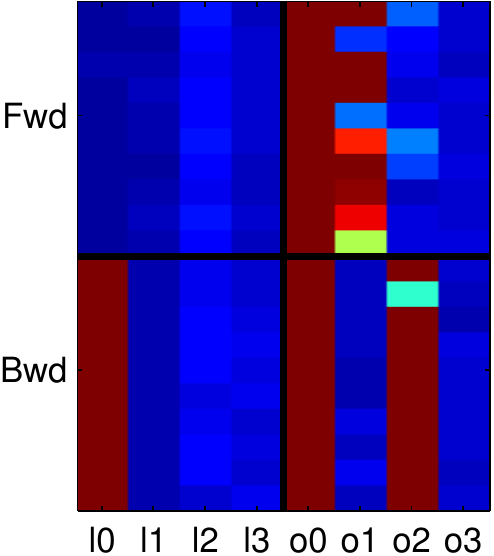}
\includegraphics[height=2.6cm]{images/results/SIM_imageLegend_crop.pdf}\\[1mm]
\caption{\textbf{Full evaluation results.} All error values for the EuRoC MAV dataset (left) and the ICL\_NUIM dataset (right): 
Each square corresponds to the (color-coded) absolute trajectory error $e_\text{ate}$ over the full sequence.
We run each of the 11 + 8 sequences (horizontal axis)
forwards (\quotes{Fwd}) and backwards (\quotes{Bwd}), 10 times each (vertical axis); 
for the EuRoC MAV dataset we further use the left and the right image stream. 
Figure~\ref{fig:MAVSIM} shows these error values aggregated as cumulative error plot (bold, continuous lines).}
\label{fig:MAVNum}
\end{figure}

\begin{figure}
\centering
ORB-SLAM:\\[0.5mm]
\includegraphics[width=0.94\linewidth]{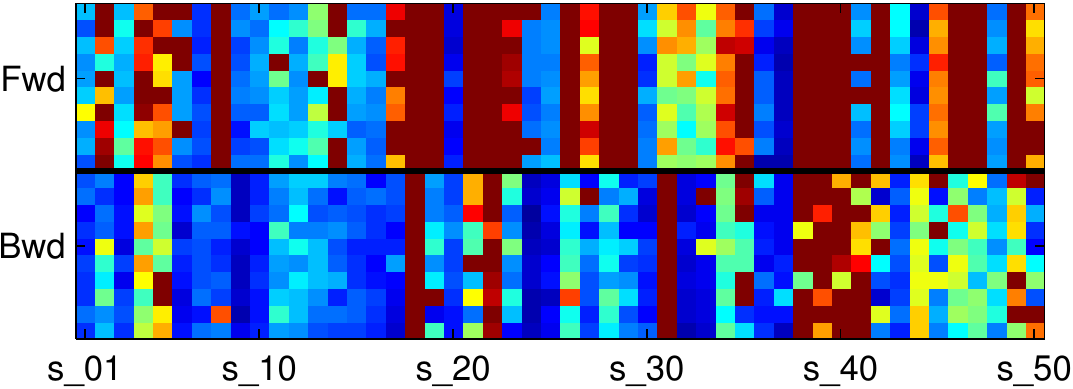}
\includegraphics[width=0.0485\linewidth]{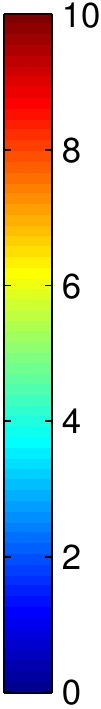}\\[1mm]
DSO:\\[0.5mm]
\includegraphics[width=0.94\linewidth]{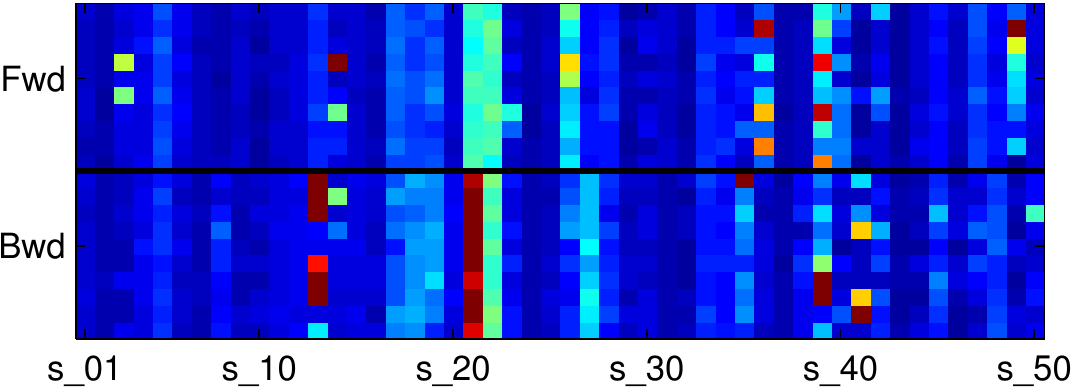}
\includegraphics[width=0.0485\linewidth]{images/results/LOOP_imageLegend_crop.pdf}\\[1mm]
\caption{\textbf{Full evaluation results.} All error values for the TUM-monoVO dataset (also see Figure \ref{fig:monoTUMdataset}). 
Each square corresponds to the (color-coded) alignment error $e_\text{align}$,
as defined in \cite{engel16archiveDataset}.
We run each of the 50 sequences (horizontal axis) forwards (\quotes{Fwd}) and backwards (\quotes{Bwd}), 10 times each (vertical axis). 
Figure~\ref{fig:accTUM} shows all these error values aggregated as cumulative error plot (bold, continuous lines).}
\label{fig:LOOPNum}
\end{figure}

	\subsection{Quantitative Comparison}
	\label{ssecQuantComp}
	Figure~\ref{fig:MAVSIM} shows the absolute trajectory RMSE $e_\text{ate}$ on the EuRoC MAV dataset and the ICL-NUIM dataset for both methods (if an algorithm gets lost within a sequence, we set $e_\text{ate} = \infty$). Figure~\ref{fig:accTUM} shows the alignment error $e_\text{align}$, as well as the rotation-drift $e_r$ and scale-drift $e_s$ for the TUM-monoVO dataset.

	In addition to the non-real-time evaluation (bold lines), we evaluate both algorithms in a hard-enforced real-time setting 
	on an Intel i7-4910MQ CPU (dashed lines).
	The direct, sparse approach clearly outperforms ORB-SLAM in accuracy and robustness both on the TUM-monoVO dataset, as well as the synthetic ICL\_NUIM dataset. 
	On the EuRoC MAV dataset, ORB-SLAM achieves a better accuracy (but lower robustness). This is due to two 
	major reasons: (1) there is no photometric calibration available, and (2) the sequences contain many small loops
	or segments where the quadrocopter \quotes{back-tracks} the way it came, allowing ORB-SLAM's local mapping component to 
	implicitly close many small and some large loops, whereas our visual odometry formulation permanently marginalizes all points 
	and frames that leave the field of view. 
	We can validate this by prohibiting ORB-SLAM from matching against any keypoints \textit{that have not been observed for more than $t_\text{max}=10$s} (lines with circle markers in Figure~\ref{fig:MAVSIM}):
	In this case, ORB-SLAM performs similar to DSO in terms of accuracy, but is less robust.
	The slight difference in robustness for DSO comes from the fact that 
	for real-time execution, tracking new frames and keyframe-creation are parallelized, thus new frames are tracked on the second-latest 
	keyframe, instead of the latest. In some rare cases -- in particular during strong exposure changes -- this causes 
	initial image alignment to fail.
	
	To show the flexibility of DSO, we include results when running at 5 times real-time speed\footnote{All images are loaded, decoded, and pinhole-rectified beforehand.}, 
	with reduced settings ($N_\mathbf{p}$=800 points, $N_f$=6 active frames, $424\!\times\!320$ image 
	resolution, $\leq\!4$ Gauss-Newton iterations after a keyframe is created): Even with such extreme settings,
	DSO achieves very good accuracy and robustness on all three datasets.
	
	Note that DSO is designed as a pure visual adometry while ORB-SLAM 
	constitutes a full SLAM system, including loop-closure detection \& correction and 
	re-localization -- all these additional abilities are neglected or switched off in this comparison.

	\subsection{Parameter Studies}
	\label{ssecParameter}
	This section aims at evaluating a number of different parameter and algorithm design choices, using the TUM-monoVO dataset.

\begin{figure}
\centering
\begin{minipage}{0.03\linewidth}\rotatebox{90}{\footnotesize ~~~number of runs}\\\end{minipage}
\begin{minipage}{.91\linewidth}\centering\includegraphics[width=1\linewidth]{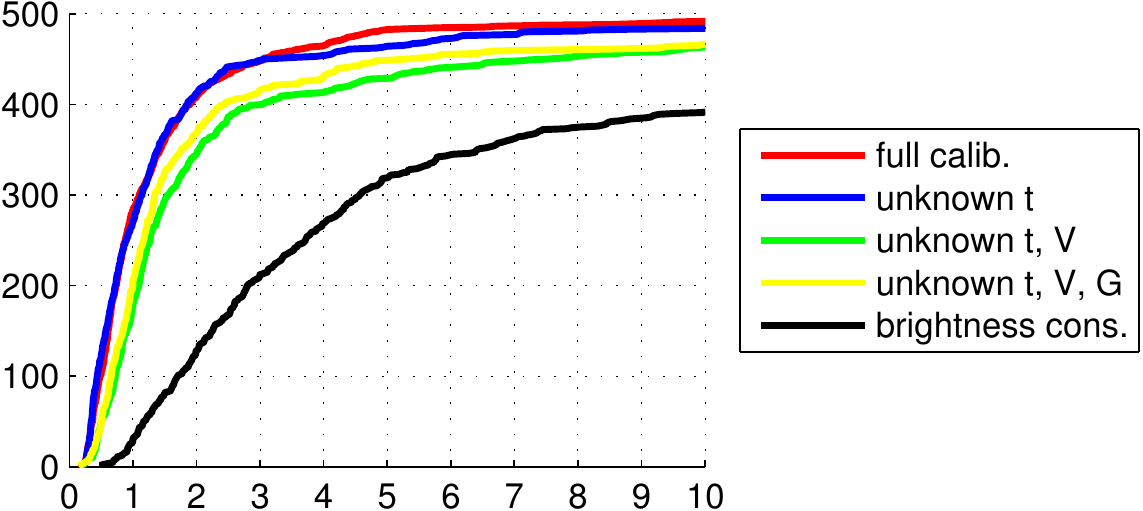}\\[-.5mm]\footnotesize  $e_\text{align}$\end{minipage}\\[2mm]
\caption{\textbf{Photometric calibration.} Errors on the TUM-monoVO dataset, when incrementally disabling photometric calibration. }
\label{fig:plotPhotoCalib}
\end{figure}

	\paragraph{Photometric Calibration.} We analyze the influence of photometric calibration, verifying that it in fact 
	increases accuracy and robustness: to this end, we incrementally disable the different components:
	\begin{enumerate}\itemsep-1mm
		\item exposure (blue): set $t_i=1$ and $\lambda_a = \lambda_b = 0$.
		\item vignette (green): set $V(\mathbf{x}) = 1$ (and 1.).
		\item response (yellow): set $G^{-1} = \text{identity}$ (and 1 -- 2.).
		\item brightness constancy (black): set $\lambda_a=\lambda_b=\infty$, i.e., disable affine brightness correction (and 1 -- 3.).
	\end{enumerate}		
	Figure~\ref{fig:plotPhotoCalib} shows the result. 
	While known exposure times seem to have little effect on the accuracy, removing vignette and response calibration does slightly decrease the 
	overall accuracy and robustness. Interestingly, only removing vignette calibration performs slightly worse than removing vignette and response calibration.
	A na\"ive brightness constancy assumption (as used in many other direct approaches like LSD-SLAM or SVO) clearly performs worst,
	since it does not account for automatic exposure changes at all.

\begin{figure}
\centering
\begin{minipage}{0.03\linewidth}\rotatebox{90}{\footnotesize ~~~number of runs}\\\end{minipage}
\begin{minipage}{.91\linewidth}\centering\includegraphics[width=1\linewidth]{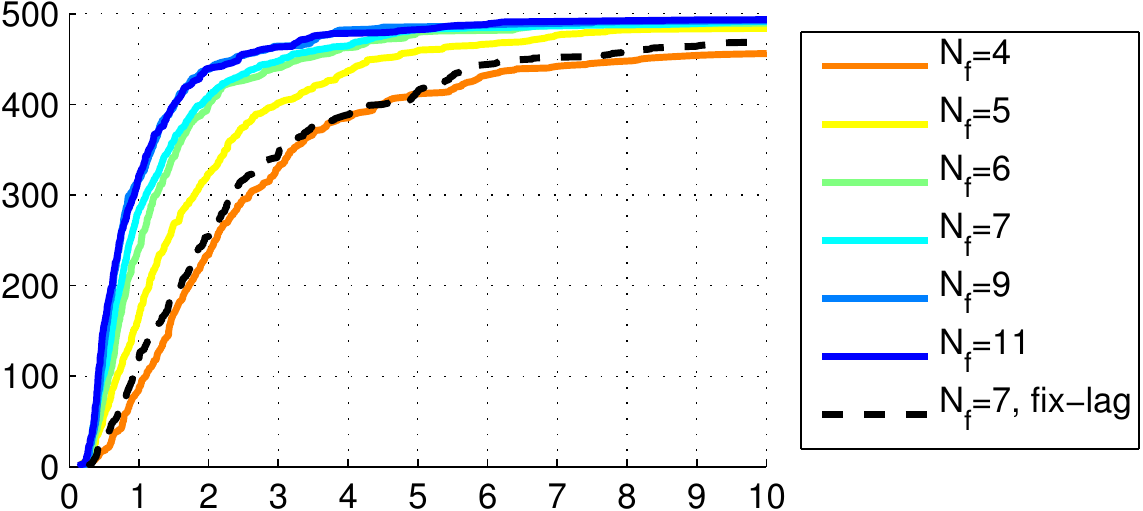}\\[-.5mm]\footnotesize  $e_\text{align}$\end{minipage}\\[2mm]
\begin{minipage}{0.03\linewidth}\rotatebox{90}{\footnotesize ~~~number of runs}\\\end{minipage}
\begin{minipage}{.91\linewidth}\centering\includegraphics[width=1\linewidth]{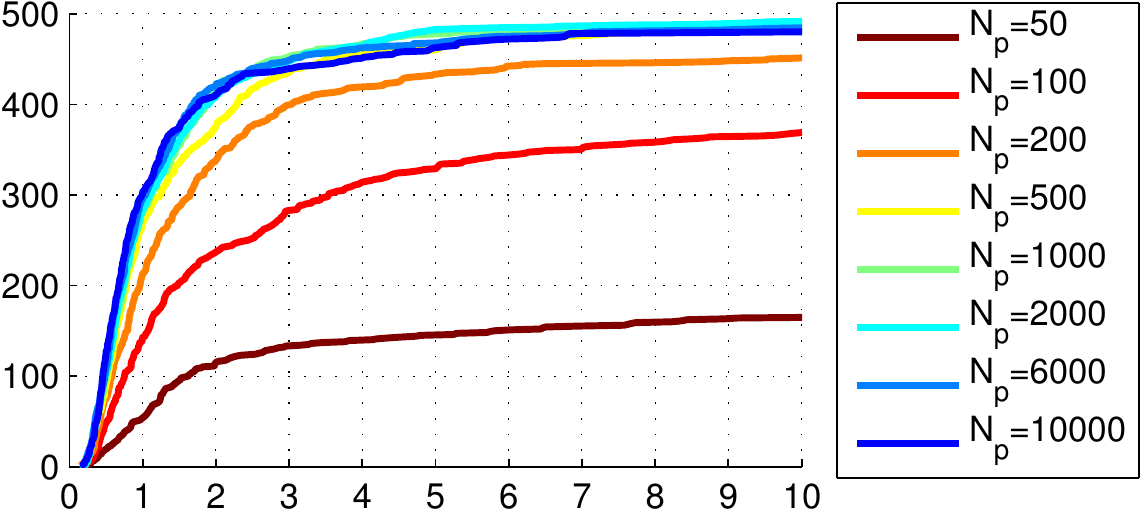}\\[-.5mm]\footnotesize  $e_\text{align}$\end{minipage}\\[2mm]
\caption{\textbf{Amount of data used.} Errors on the TUM-monoVO dataset, when changing the size of the optimization window (top)
and the number of points (bottom). Using more than $N_p=500$ points or $N_f=7$ active frames has only marginal impact. Note that as real-time default setting, 
we use $N_p=2000$ and $N_f=7$, mainly to obtain denser reconstructions.}
\label{fig:dataAmount}
\end{figure}	
	
	\paragraph{Amount of Data.} We analyze the effect of changing the amount of data used, by varying
	the number of active points $N_p$, as well as the number of frames in the active window $N_f$. Note that increasing $N_f$ allows to keep 
	more observations per point: For any point we only ever keep observations in active frames; thus the number of 
	observations when marginalizing a point is limited to $N_f$ (see Section~\ref{ssecOptimization}). 
	Figure~\ref{fig:dataAmount} summarizes the result.
	We can observe that the benefit of simply using \textit{more} data quickly flattens off after $N_p=500$ points. 
	At the same time, the number of active frames has little influence after $N_f=7$, while increasing the runtime quadratically. 
	We further evaluate a fixed-lag marginalization strategy (i.e., always marginalize the oldest keyframe, instead of using the proposed
	distancescore) as in \cite{leutenegger2015ijrr}: this performs significantly worse.

\begin{figure}
\centering
\begin{minipage}{0.03\linewidth}\rotatebox{90}{\footnotesize ~~~number of runs}\\\end{minipage}
\begin{minipage}{.455\linewidth}\centering\includegraphics[width=1\linewidth]{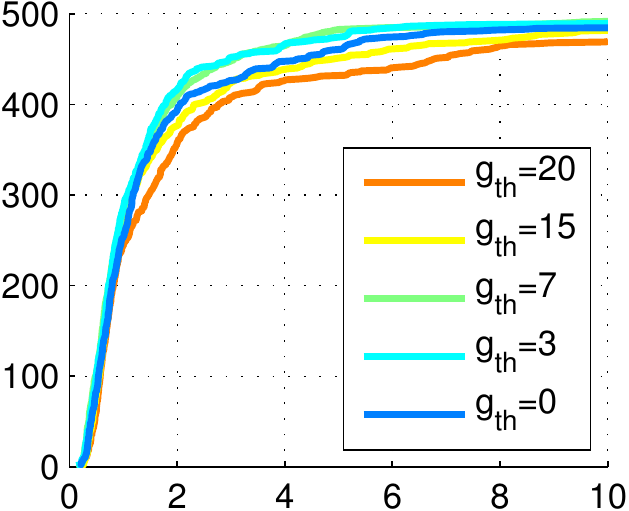}\\[-.5mm]\footnotesize  $e_\text{align}$\end{minipage}
\begin{minipage}{.455\linewidth}\centering\includegraphics[width=1\linewidth]{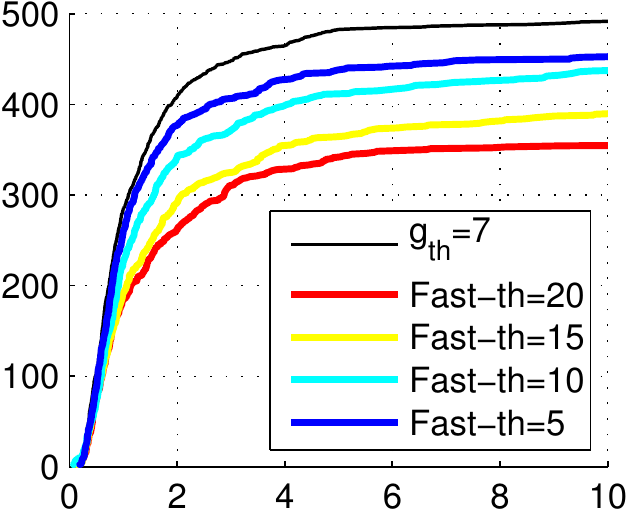}\\[-.5mm]\footnotesize  $e_\text{align}$\end{minipage}\\[2mm]
\caption{\textbf{Selection of data used.} Errors on the TUM-monoVO dataset, when changing the \textit{type} of data used. Left:
Errors for different gradient thresholds $g_\text{th}$, which seems to have a limited impact on the algorithms accuracy. 
Right: Errors when only using FAST corners, at different thresholds. Using only FAST corners significantly reduces accuracy and
robustness, showing that the ability to use data from edges and weakly textured surfaces does have a real benefit.}
\label{fig:dataSelection}
\end{figure}

	\paragraph{Selection of Data.} In addition to evaluating the effect of the number of residuals used, it is interesting to look at
	which data is used -- in particular since one of the main benefits of
	a direct approach is the ability to sample from all points, instead of only using corners. 
	To this end, we vary the gradient threshold for point selection, $g_\text{th}$; the result is summarized
	 in Figure~\ref{fig:dataSelection}. While there seems to be a sweet spot around $g_\text{th}=7$
	(if $g_\text{th}$ is too large, for some scenes not enough well-distributed points are available to sample 
	from -- if it is too low, too much weight will be given to data with a low signal-to-noise ratio), 
	the overall impact is relatively low.
	
	More interestingly, we analyse the effect of \textit{only using corners}, by restricting point candidates
	to FAST corners only. We can clearly see that only using corners 
	significantly decreases performance. Note that for lower FAST thresholds, many false \quotes{corners} will be
	detected along edges, which our method can still use, in contrast to 
	indirect methods for which such points will be outliers. 
	In fact, ORB-SLAM achieves best performance using the default threshold of 20.

\begin{figure}
\centering
\begin{minipage}{0.03\linewidth}\rotatebox{90}{\footnotesize ~~~number of runs}\\\end{minipage}
\begin{minipage}{.71\linewidth}\centering\includegraphics[width=1\linewidth]{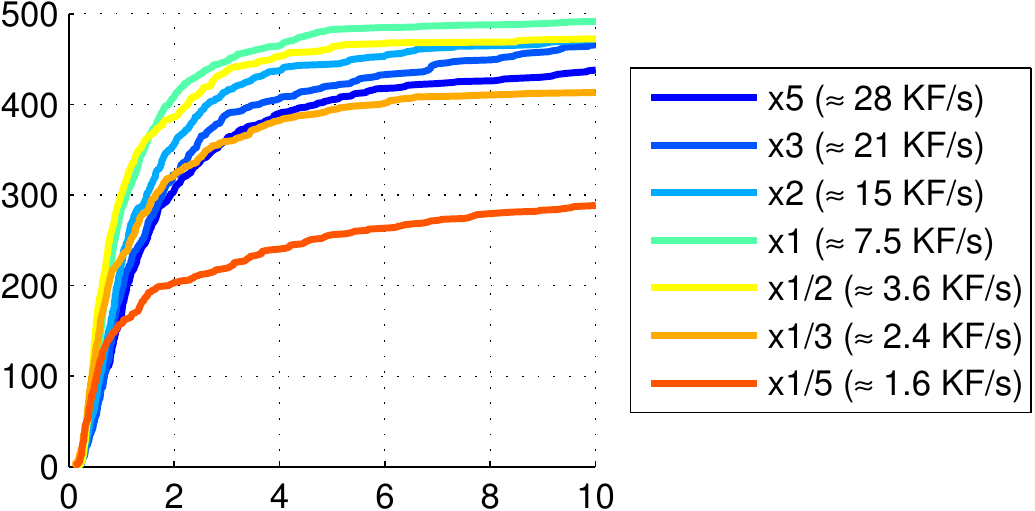}\\[-.5mm]\footnotesize  $e_\text{align}$\end{minipage}\\[2mm]
\caption{\textbf{Number of keyframes.} Errors on the TUM-monoVO dataset, when changing the number of keyframes 
taken via the threshold $T_\text{kf}$.}
\label{fig:numKF}
\end{figure}

	\paragraph{Number of Keyframes.} We analyze the number of keyframes taken by varying $T_\text{kf}$ (see Section~\ref{sssecFrameManagement}). 
	For each value of $T_\text{kf}$ we give the resulting average number of keyframes per second; the default 
	setting $T_\text{kf}=1$ results in 8 keyframes per second, which is easily achieved in real time.
	The result is summarized in Figure~\ref{fig:numKF}.
	Taking too few keyframes (less than 4 per second) reduces the robustness, mainly
	in situations with strong occlusions / dis-occlusions, e.g., when walking through doors. 
	Taking too many keyframes, on the other hand (more than 15 per second), decreases accuracy.
	This is because taking more keyframes causes them to be marginalized earlier (since $N_f$ is fixed),
	thereby accumulating linearizations around earlier (and less accurate) linearization points.

\begin{figure}
\centering
\begin{minipage}{0.03\linewidth}\rotatebox{90}{\footnotesize ~~~number of runs}\end{minipage}
\begin{minipage}{.55\linewidth}\centering\includegraphics[width=1\linewidth]{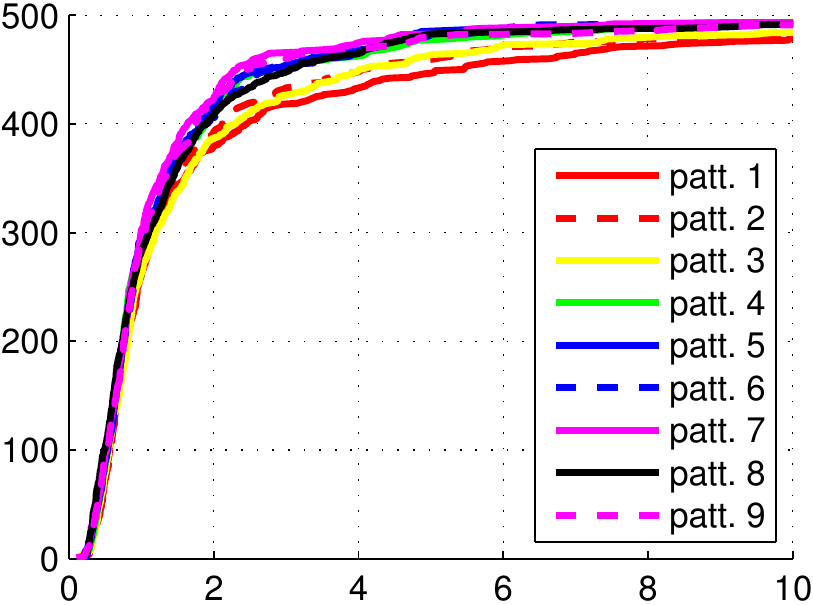}\\[-.5mm]\footnotesize  $e_\text{align}$\end{minipage}
\begin{minipage}{.35\linewidth}

	\begin{tikzpicture}[scale=0.8]
	\fill[blue!80!white] (0.4,0.4) rectangle (0.6,0.6);
	\fill[blue!40!white] (0.4,0.4)++(0.2,0) rectangle +(0.2,0.2);
	\fill[blue!40!white] (0.4,0.4)++(-0.2,0) rectangle +(0.2,0.2);
	\fill[blue!40!white] (0.4,0.4)++(0,0.2) rectangle +(0.2,0.2);
	\fill[blue!40!white] (0.4,0.4)++(0,-0.2) rectangle +(0.2,0.2);
	\draw[step=.2cm,gray,thin] (0.1,0.1) grid (0.9,0.9);	
	\draw (0.5,0.2) node[anchor=north] {\tiny patt. 1};
	\end{tikzpicture}
	\begin{tikzpicture}[scale=0.8]
	\fill[blue!80!white] (0.4,0.4) rectangle (0.6,0.6);
	\fill[blue!40!white] (0.4,0.4)++(0.2,0.2) rectangle +(0.2,0.2);
	\fill[blue!40!white] (0.4,0.4)++(-0.2,0.2) rectangle +(0.2,0.2);
	\fill[blue!40!white] (0.4,0.4)++(0.2,-0.2) rectangle +(0.2,0.2);
	\fill[blue!40!white] (0.4,0.4)++(-0.2,-0.2) rectangle +(0.2,0.2);
	\draw[step=.2cm,gray,thin] (0.1,0.1) grid (0.9,0.9);	
	\draw (0.5,0.2) node[anchor=north] {\tiny patt. 2};
	\end{tikzpicture}
	\begin{tikzpicture}[scale=0.8]
	\fill[blue!80!white] (0.4,0.4) rectangle (0.6,0.6);
	\fill[blue!40!white] (0.4,0.4)++(0.2,0.2) rectangle +(0.2,0.2);
	\fill[blue!40!white] (0.4,0.4)++(-0.2,0.2) rectangle +(0.2,0.2);
	\fill[blue!40!white] (0.4,0.4)++(0.2,-0.2) rectangle +(0.2,0.2);
	\fill[blue!40!white] (0.4,0.4)++(-0.2,-0.2) rectangle +(0.2,0.2);
	\fill[blue!40!white] (0.4,0.4)++(0.2,0) rectangle +(0.2,0.2);
	\fill[blue!40!white] (0.4,0.4)++(-0.2,0) rectangle +(0.2,0.2);
	\fill[blue!40!white] (0.4,0.4)++(0,0.2) rectangle +(0.2,0.2);
	\fill[blue!40!white] (0.4,0.4)++(0,-0.2) rectangle +(0.2,0.2);
	\draw[step=.2cm,gray,thin] (0.1,0.1) grid (0.9,0.9);	
	\draw (0.5,0.2) node[anchor=north] {\tiny patt. 3};
	\end{tikzpicture}
	\begin{tikzpicture}[scale=0.6]
	\fill[blue!80!white] (0.4,0.4) rectangle (0.6,0.6);
	\fill[blue!40!white] (0.4,0.4)++(0.4,0) rectangle +(0.2,0.2);
	\fill[blue!40!white] (0.4,0.4)++(-0.4,0) rectangle +(0.2,0.2);
	\fill[blue!40!white] (0.4,0.4)++(0,0.4) rectangle +(0.2,0.2);
	\fill[blue!40!white] (0.4,0.4)++(0,-0.4) rectangle +(0.2,0.2);
	\fill[blue!40!white] (0.4,0.4)++(0.2,0.2) rectangle +(0.2,0.2);
	\fill[blue!40!white] (0.4,0.4)++(-0.2,0.2) rectangle +(0.2,0.2);
	\fill[blue!40!white] (0.4,0.4)++(0.2,-0.2) rectangle +(0.2,0.2);
	\fill[blue!40!white] (0.4,0.4)++(-0.2,-0.2) rectangle +(0.2,0.2);
	\draw[step=.2cm,gray,thin] (-0.1,-0.1) grid (1.1,1.1);	
	\draw (0.5,0) node[anchor=north] {\tiny patt. 4};
	\end{tikzpicture}
	\begin{tikzpicture}[scale=0.6]
	\fill[blue!80!white] (0.4,0.4) rectangle (0.6,0.6);
	\fill[blue!40!white] (0.4,0.4)++(0.4,0.4) rectangle +(0.2,0.2);
	\fill[blue!40!white] (0.4,0.4)++(-0.4,0.4) rectangle +(0.2,0.2);
	\fill[blue!40!white] (0.4,0.4)++(0.4,-0.4) rectangle +(0.2,0.2);
	\fill[blue!40!white] (0.4,0.4)++(-0.4,-0.4) rectangle +(0.2,0.2);
	\fill[blue!40!white] (0.4,0.4)++(0.4,0) rectangle +(0.2,0.2);
	\fill[blue!40!white] (0.4,0.4)++(-0.4,0) rectangle +(0.2,0.2);
	\fill[blue!40!white] (0.4,0.4)++(0,0.4) rectangle +(0.2,0.2);
	\fill[blue!40!white] (0.4,0.4)++(0,-0.4) rectangle +(0.2,0.2);
	\fill[blue!40!white] (0.4,0.4)++(0.2,0.2) rectangle +(0.2,0.2);
	\fill[blue!40!white] (0.4,0.4)++(-0.2,0.2) rectangle +(0.2,0.2);
	\fill[blue!40!white] (0.4,0.4)++(0.2,-0.2) rectangle +(0.2,0.2);
	\fill[blue!40!white] (0.4,0.4)++(-0.2,-0.2) rectangle +(0.2,0.2);
	\draw[step=.2cm,gray,thin] (-0.1,-0.1) grid (1.1,1.1);	
	\draw (0.5,0) node[anchor=north] {\tiny patt. 5};
	\end{tikzpicture}
	\begin{tikzpicture}[scale=0.6]
	\fill[blue!40!white] (0,0) rectangle (1,1);
	\fill[blue!80!white] (0.4,0.4) rectangle (0.6,0.6);
	\draw[step=.2cm,gray,thin] (-0.1,-0.1) grid (1.1,1.1);	
	\draw (0.5,0) node[anchor=north] {\tiny patt. 6};
	\end{tikzpicture}\hfill
	\begin{tikzpicture}[scale=0.6]
	\fill[blue!80!white] (0.4,0.4) rectangle (0.6,0.6);
	\fill[blue!40!white] (0.4,0.4)++(0.4,0) rectangle +(0.2,0.2);
	\fill[blue!40!white] (0.4,0.4)++(-0.4,0) rectangle +(0.2,0.2);
	\fill[blue!40!white] (0.4,0.4)++(0,0.4) rectangle +(0.2,0.2);
	\fill[blue!40!white] (0.4,0.4)++(0,-0.4) rectangle +(0.2,0.2);
	\fill[blue!40!white] (0.4,0.4)++(0.2,0.2) rectangle +(0.2,0.2);
	\fill[blue!40!white] (0.4,0.4)++(-0.2,0.2) rectangle +(0.2,0.2);
	\fill[blue!40!white] (0.4,0.4)++(-0.2,-0.2) rectangle +(0.2,0.2);
	\draw[step=.2cm,gray,thin] (-0.1,-0.1) grid (1.1,1.1);	
	\draw (0.5,0) node[anchor=north] {\tiny patt. 8};
	\end{tikzpicture}\hfill
	\begin{tikzpicture}[scale=0.5]
	\fill[blue!80!white] (0.4,0.4) rectangle (0.6,0.6);
	\fill[blue!40!white] (0.4,0.4)++(0.4,0.4) rectangle +(0.2,0.2);
	\fill[blue!40!white] (0.4,0.4)++(-0.4,0.4) rectangle +(0.2,0.2);
	\fill[blue!40!white] (0.4,0.4)++(0.4,-0.4) rectangle +(0.2,0.2);
	\fill[blue!40!white] (0.4,0.4)++(-0.4,-0.4) rectangle +(0.2,0.2);
	\fill[blue!40!white] (0.4,0.4)++(0.4,0) rectangle +(0.2,0.2);
	\fill[blue!40!white] (0.4,0.4)++(-0.4,0) rectangle +(0.2,0.2);
	\fill[blue!40!white] (0.4,0.4)++(0,0.4) rectangle +(0.2,0.2);
	\fill[blue!40!white] (0.4,0.4)++(0,-0.4) rectangle +(0.2,0.2);
	\fill[blue!40!white] (0.4,0.4)++(0.2,0.2) rectangle +(0.2,0.2);
	\fill[blue!40!white] (0.4,0.4)++(-0.2,0.2) rectangle +(0.2,0.2);
	\fill[blue!40!white] (0.4,0.4)++(0.2,-0.2) rectangle +(0.2,0.2);
	\fill[blue!40!white] (0.4,0.4)++(-0.2,-0.2) rectangle +(0.2,0.2);
	\fill[blue!40!white] (0.4,0.4)++(-0.6,-0.2) rectangle +(0.2,0.2);	
	\fill[blue!40!white] (0.4,0.4)++(-0.6,0.2) rectangle +(0.2,0.2);	
	\fill[blue!40!white] (0.4,0.4)++(0.6,-0.2) rectangle +(0.2,0.2);	
	\fill[blue!40!white] (0.4,0.4)++(0.6,0.2) rectangle +(0.2,0.2);	
	\fill[blue!40!white] (0.4,0.4)++(-0.2,-0.6) rectangle +(0.2,0.2);	
	\fill[blue!40!white] (0.4,0.4)++(0.2,-0.6) rectangle +(0.2,0.2);	
	\fill[blue!40!white] (0.4,0.4)++(-0.2,0.6) rectangle +(0.2,0.2);	
	\fill[blue!40!white] (0.4,0.4)++(0.2,0.6) rectangle +(0.2,0.2);	
	\draw[step=.2cm,gray,thin] (-0.3,-0.3) grid (1.3,1.3);	
	\draw (0.5,-0.2) node[anchor=north] {\tiny patt. 7};
	\end{tikzpicture}
	\begin{tikzpicture}[scale=0.4]
	\foreach \x in {-2,...,2}
    		\foreach \y in {-2,...,2} 
			{\fill[blue!40!white] (0.4,0.4)++(0.4*\x,0.4*\y) rectangle +(0.2,0.2);}
	\fill[blue!80!white] (0.4,0.4) rectangle (0.6,0.6);
	\draw[step=.2cm,gray,thin] (-0.5,-0.5) grid (1.5,1.5);	
	\draw (0.5,-0.4) node[anchor=north] {\tiny patt. 9};
	\end{tikzpicture}
\end{minipage}\\[2mm]
\caption{\textbf{Residual pattern.} Errors on the TUM-monoVO dataset for some of the evaluated patterns $\mathcal{N}_\mathbf{p}$. Using only a $3 \times 3$ neighborhood
seems to perform slightly worse -- using more than the proposed 8-pixel pattern however seems to have little benefit -- at the same time, using a larger neighbourhood 
increases the computational demands. Note that these results may vary with low-level properties of the used camera and lens, such as the point spread function.}
\label{fig:patternEval}
\end{figure}
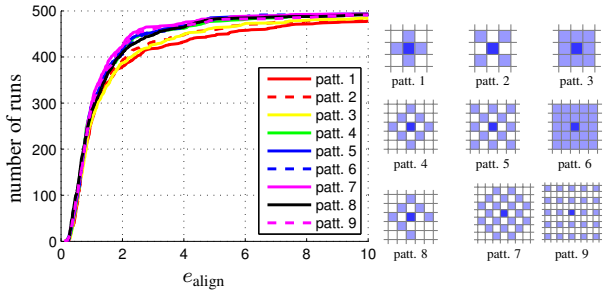

	\paragraph{Residual Pattern.} 
	We test different residual patterns for $\mathcal{N}_\mathbf{p}$, covering smaller or larger areas. The result is 
	shown in Figure~\ref{fig:patternEval}.

	\subsection{Geometric vs. Photometric Noise Study}
	\label{ssecResultsNoise}
	The fundamental difference between the proposed direct model and the indirect model is the noise assumption. 
	\textit{The direct approach models photometric noise}, i.e., additive noise on pixel intensities. 
	In contrast, \textit{the indirect approaches models geometric noise}, i.e., additive noise on the
	$(u,v)$-position of a point in the image plane, assuming that keypoint descriptors are robust to photometric noise.
	It therefore comes at no surprise that the indirect approach is significantly more robust to geometric noise in the data.
	In turn, the direct approach performs better in the presence of strong photometric noise -- which keypoint-descriptors 
	(operating on a purely local level) fail to filter out.
	We verify this by analyzing tracking accuracy on the TUM-monoVO dataset, when artificially adding (a) geometric noise, 
	and (b) photometric noise to the images.

\begin{figure}
\includegraphics[width=.245\linewidth]{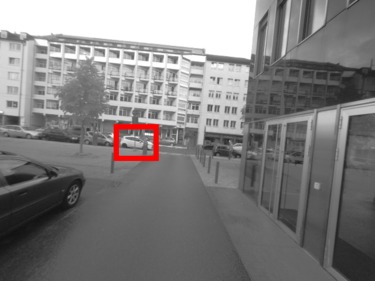}\hspace{-.5mm}
\raisebox{1pt}{{\setlength{\fboxsep}{0pt}\setlength{\fboxrule}{1pt}\fcolorbox{red}{red}{\includegraphics[width=.23\linewidth]{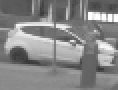}}}}\hspace{-.5mm}
\includegraphics[width=.245\linewidth]{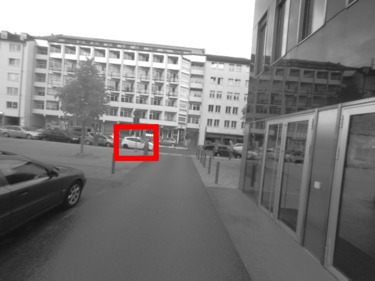}\hspace{-.5mm}
\raisebox{1pt}{{\setlength{\fboxsep}{0pt}\setlength{\fboxrule}{1pt}\fcolorbox{red}{red}{\includegraphics[width=.23\linewidth]{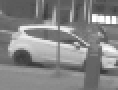}}}}\\
\footnotesize 
\hspace*{0.9cm} $I$ 
\hspace*{1.1cm} $I$ (close-up) 
\hspace*{1.2cm} $I'_g$ 
\hspace*{1.0cm} $I'_g$ (close-up)\\[1mm]
\begin{minipage}{0.03\linewidth}\rotatebox{90}{\footnotesize number of runs}\\\end{minipage}
\begin{minipage}{0.95\linewidth}
\includegraphics[height=2.7cm]{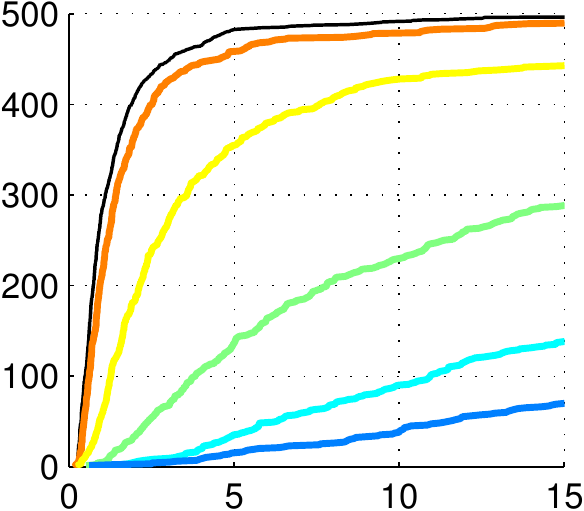}
\includegraphics[height=2.7cm]{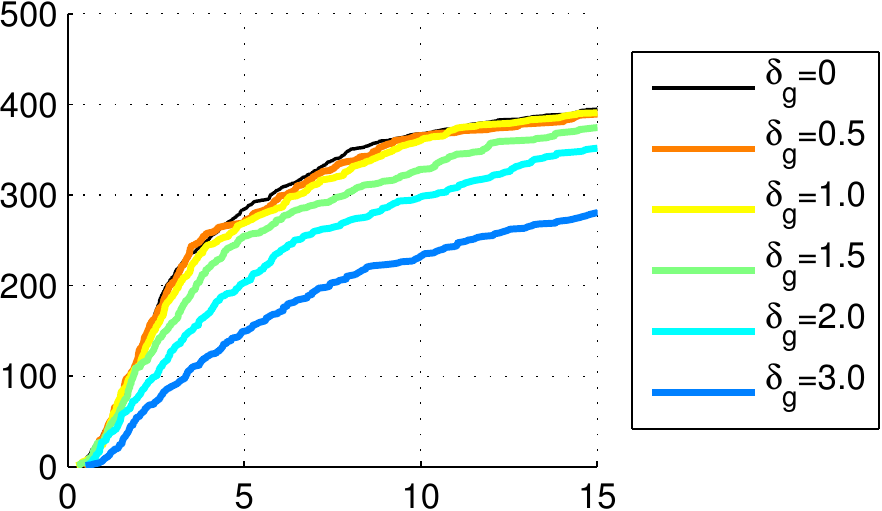}\\
\footnotesize  \hspace*{0.8cm} $e_\text{align}$ (DSO) \hspace{1.0cm} $e_\text{align}$ (ORB-SLAM)
\end{minipage}\\
\caption{\textbf{Geometric noise.} Effect of applying low-frequency geometric noise to the image,
simulating geometric distortions such as a rolling shutter (evaluated on the TUM-monoVO dataset). The top row shows an example image with $\delta_g=2$.
While the effect is hardly visible to the human eye (observe that the close-up is slightly shifted), it has a severe 
impact on SLAM accuracy, in particular when using a direct model. Note that the  distortion caused by a standard rolling shutter camera easily surpasses $\delta_g=3$.}
\label{fig:geoNoise}
\end{figure}

	\paragraph{Geometric Noise.} For each frame, we separately generate a low-frequency random flow-map $N_g \colon \Omega \to \mathbb{R}^2$ 
	by up-sampling a $3\!\times\!3$ grid filled with uniformly distributed random values from
	$[-\delta_g, \delta_g]^2$ (using bicubic interpolation). 
	We then perturb the original image by shifting each pixel $\mathbf{x}$ by $N_g(\mathbf{x})$:
	\begin{align}
		I_g'(\mathbf{x}) := I(\mathbf{x}+N_g(\mathbf{x})).
	\end{align}
	This procedure simulates noise originating from (unmodeled) rolling shutter or inaccurate geometric camera calibration.
	Figure~\ref{fig:geoNoise} visualizes an example of the resulting noise pattern, as well as the accuracy of ORB-SLAM and DSO for different values of 
	$\delta_g$. As expected, we can clearly observe how DSO's performance 
	quickly deteriorates with added geometric noise, whereas ORB-SLAM is much less affected. 
	This is because the first step in the indirect pipeline -- keypoint detection and extraction -- is not affected
	by low-frequency geometric noise, as it operates on a purely local level. 
	The second step then optimizes a geometric noise model -- which not surprisingly deals well with geometric noise. 
	In the direct approach, in turn, geometric noise is not modeled, and thus has a much more severe effect -- in fact,
	for $\delta_g > 1.5$ there likely exists no state for which all residuals are within the validity radius of the 
	linearization of $I$; thus optimization fails entirely (which can be alleviated by using a coarser pyramid level).
	Note that this result also suggests that the proposed direct model is more susceptible to inaccurate 
	intrinsic camera calibration than the indirect approach -- in turn, it may benefit more from accurate, 
	non-parametric intrinsic calibration.

\begin{figure}
\includegraphics[width=.245\linewidth]{images/photoGeoError/imgRawB.jpg}\hspace{-.5mm}
\raisebox{1pt}{{\setlength{\fboxsep}{0pt}\setlength{\fboxrule}{1pt}\fcolorbox{red}{red}{\includegraphics[width=.23\linewidth]{images/photoGeoError/imgRaw_cu.png}}}}\hspace{-.5mm}
\includegraphics[width=.245\linewidth]{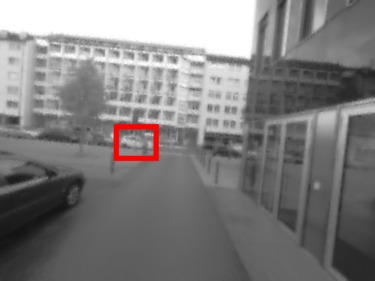}\hspace{-.5mm}
\raisebox{1pt}{{\setlength{\fboxsep}{0pt}\setlength{\fboxrule}{1pt}\fcolorbox{red}{red}{\includegraphics[width=.23\linewidth]{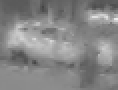}}}}\\
\footnotesize 
\hspace*{0.9cm} $I$ 
\hspace*{1.1cm} $I$ (close-up) 
\hspace*{1.2cm} $I'_p$ 
\hspace*{1.0cm} $I'_p$ (close-up)\\[1mm]
\begin{minipage}{0.03\linewidth}\rotatebox{90}{\footnotesize number of runs}\\\end{minipage}
\begin{minipage}{0.95\linewidth}
\includegraphics[height=2.7cm]{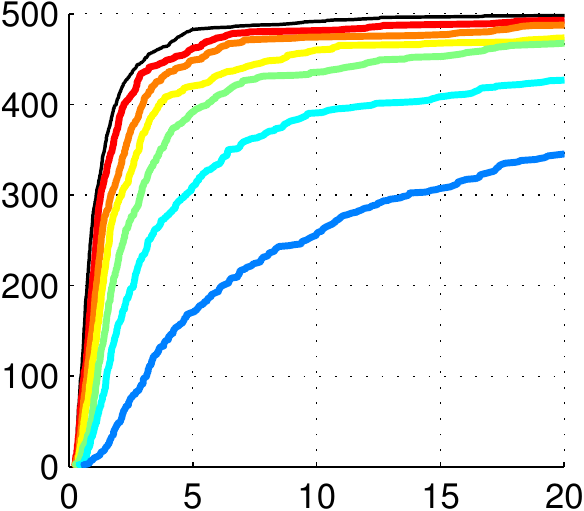}
\includegraphics[height=2.7cm]{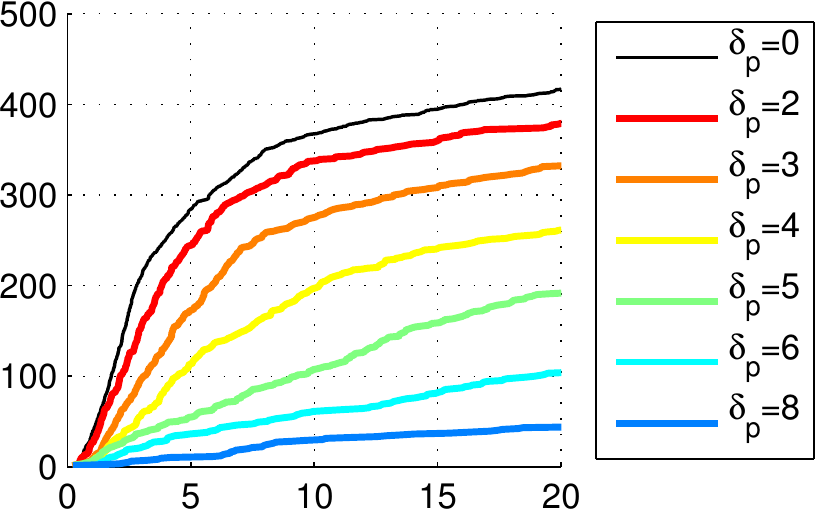}\\
\footnotesize  \hspace*{0.8cm} $e_\text{align}$ (DSO) \hspace{1.0cm} $e_\text{align}$ (ORB-SLAM)
\end{minipage}\\
\caption{\textbf{Photometric noise.} Effect of applying high-frequent, non-isotropic blur to the image,
simulating photometric noise (evaluated on the TUM-monoVO dataset). The top row shows an example image with $\delta_p=6$, the effect is clearly visible.
Since the direct approach models a photometric error, it is more robust to this type of noise than indirect methods.}
\label{fig:photoNoise}
\end{figure}

	\paragraph{Photometric Noise.} For each frame, we separately generate a high-frequency random blur-map $N_p \colon \Omega \to \mathbb{R}^2$ 
	by up-sampling a $300\!\times\!300$ grid filled with uniformly distributed random values in $[-\delta_p, \delta_p]^2$. 
	We then perturb the original image by adding anisotropic blur with standard deviation $N_p(\mathbf{x})$ to pixel $\mathbf{x}$:
	\begin{align}
		I_p'(\mathbf{x}) := \int_{\mathbb{R}^2} \phi(\boldsymbol\delta; N_p(\mathbf{x})^2) I(\mathbf{x}+\boldsymbol\delta) ~~ d \boldsymbol\delta,
	\end{align}
	where $\phi(\cdot; N_p(\mathbf{x})^2)$ denotes a 2D Gaussian kernel with standard deviation $N_p(\mathbf{x})$.
	Figure~\ref{fig:photoNoise} shows the result. We can observe that DSO is slightly more robust to photometric noise 
	than ORB-SLAM -- this is because (purely local) keypoint matching fails for high photometric noise, whereas a joint 
	optimization of the photometric error better overcomes the introduced distortions.\\[-3mm]

	To summarize: While the direct approach outperforms the indirect approach on well-calibrated data, it is
	ill-suited in the presence of strong geometric noise, e.g., originating from a rolling shutter or inaccurate intrinsic calibration.
	In practice, this makes the indirect model superior for smartphones or off-the-shelf webcams, since these 
	were designed to capture videos for human consumption -- prioritizing resolution and light-sensitivity over 
	geometric precision. 
	In turn, the direct approach offers superior performance on data captured with dedicated cameras for machine-vision, 
	since these put more importance on geometric precision, rather than capturing appealing images for human consumption.
	Note that this can be resolved by tightly integrating the rolling shutter into the model, as done, e.g., in \cite{lovegrove2013spline,li2013icra,kerl15iccv}.

	\vspace*{-1.5mm}\subsection{Qualitative Results}
\begin{figure}
\centering
\includegraphics[width=.36\linewidth]{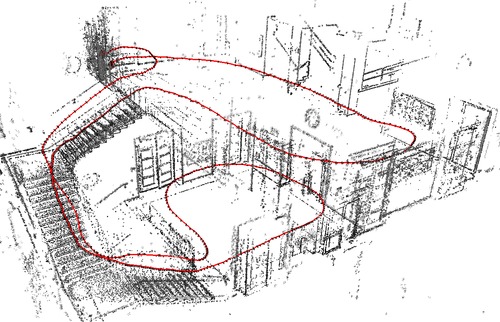}\hspace{-0.5mm}
\includegraphics[width=.305\linewidth]{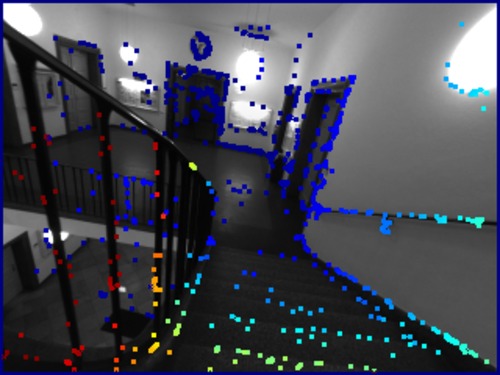}\hspace{-0.5mm}
\includegraphics[width=.305\linewidth]{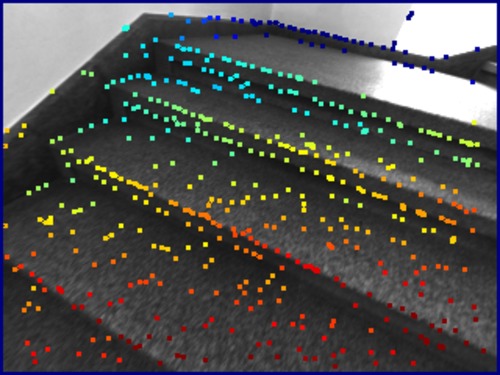}\\
\includegraphics[width=.36\linewidth]{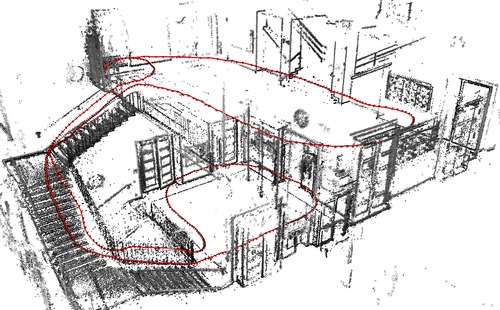}\hspace{-0.5mm}
\includegraphics[width=.305\linewidth]{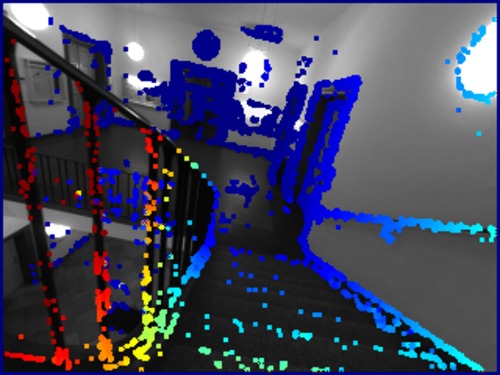}\hspace{-0.5mm}
\includegraphics[width=.305\linewidth]{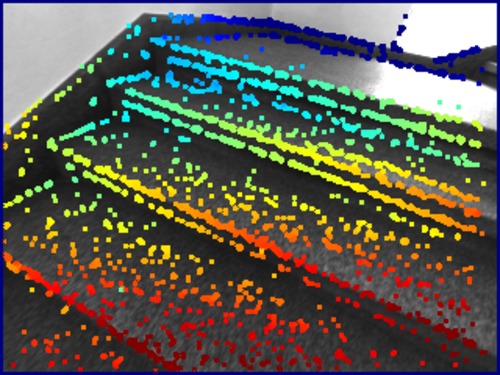}\\
\includegraphics[width=.36\linewidth]{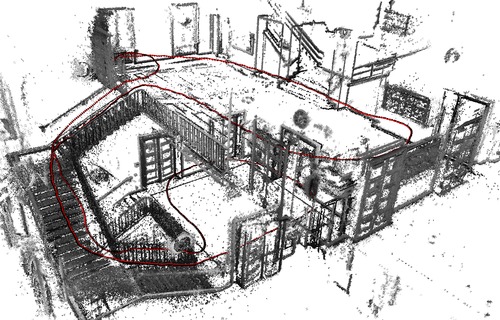}\hspace{-0.5mm}
\includegraphics[width=.305\linewidth]{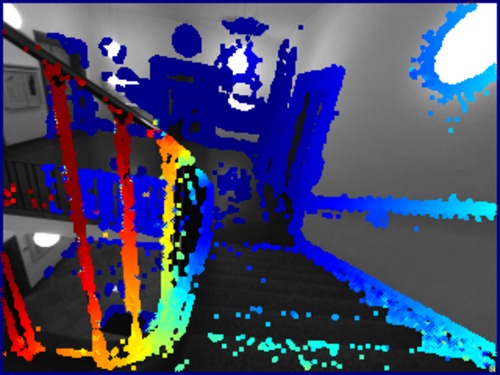}\hspace{-0.5mm}
\includegraphics[width=.305\linewidth]{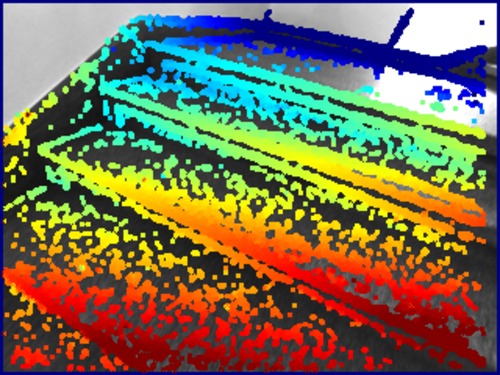}\\
\caption{\textbf{Point density.} 3D point cloud and some coarse depth maps, i.e., the most recent keyframe 
with all $N_p$ active points projected into it) for $N_p$=500 (top), $N_p$=2000 (middle), 
and $N_p$=10000 (bottom). }\vspace{-1mm}
\label{fig:densityComparison}
\end{figure}

\vspace*{-0.5mm}
	\label{secResultsQual}
	In addition to accurate camera tracking, DSO computes 3D points on all gradient-rich areas,
	including edges -- resulting in point-cloud reconstructions similar to the semi-dense reconstructions of LSD-SLAM.
	The density then directly corresponds to how many points we keep in the active window $N_p$. 
	Figure~\ref{fig:densityComparison} shows some examples.
	
\begin{figure*}
\centering
\includegraphics[width=.99\linewidth]{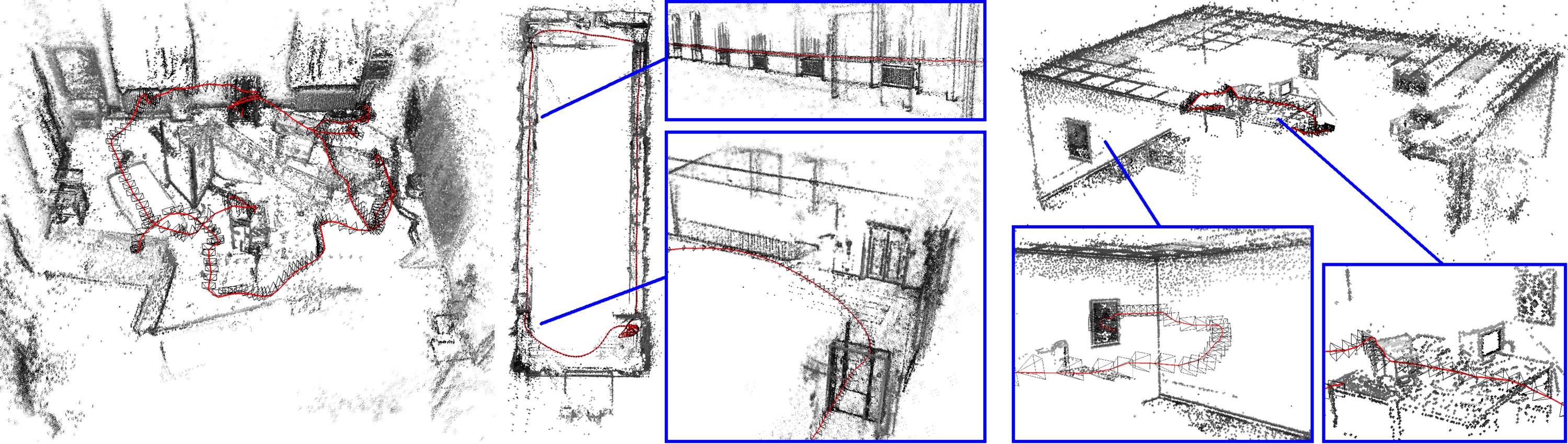}\\
\includegraphics[width=.163\linewidth]{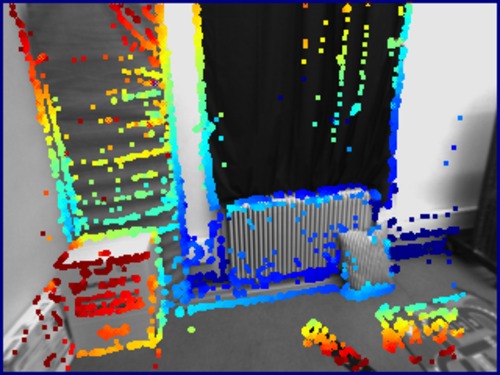}\hspace{-0.5mm}
\includegraphics[width=.163\linewidth]{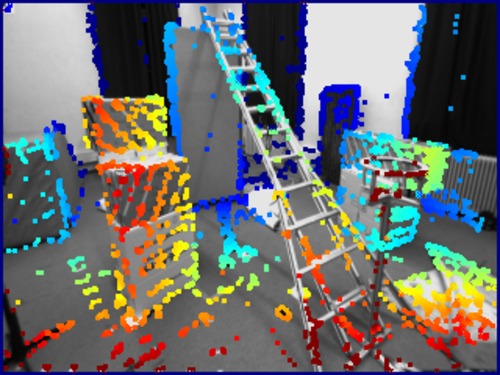}\hfill
\includegraphics[width=.163\linewidth]{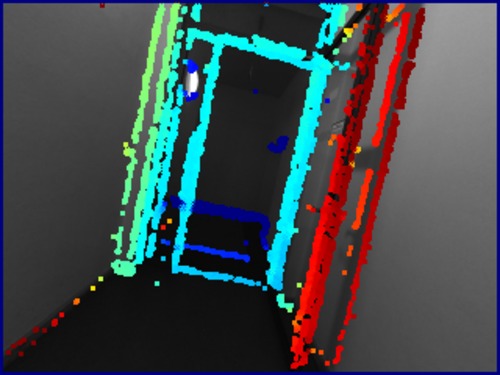}\hspace{-0.5mm}
\includegraphics[width=.163\linewidth]{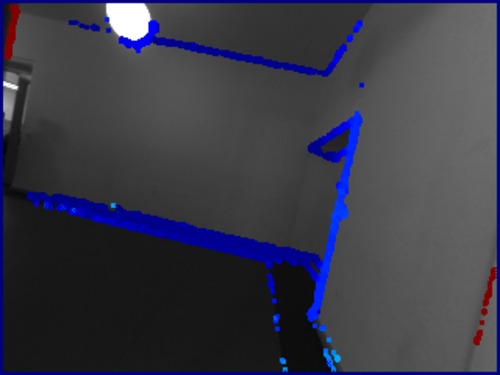}\hfill
\includegraphics[width=.163\linewidth]{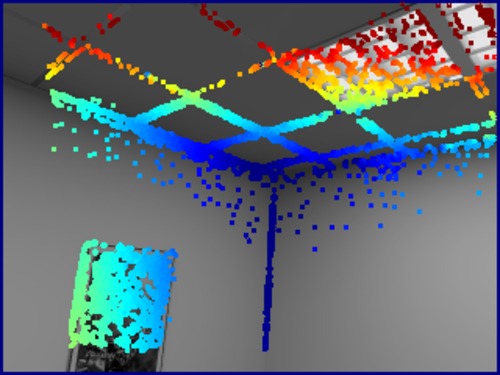}\hspace{-0.5mm}
\includegraphics[width=.163\linewidth]{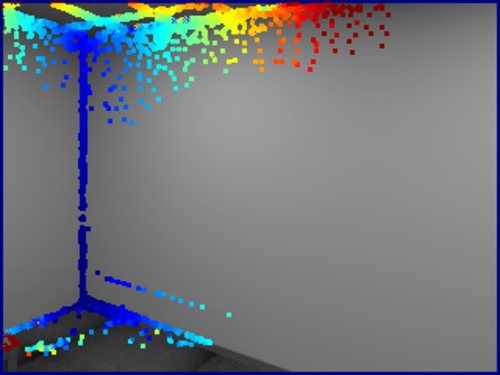}\\
\includegraphics[width=.163\linewidth]{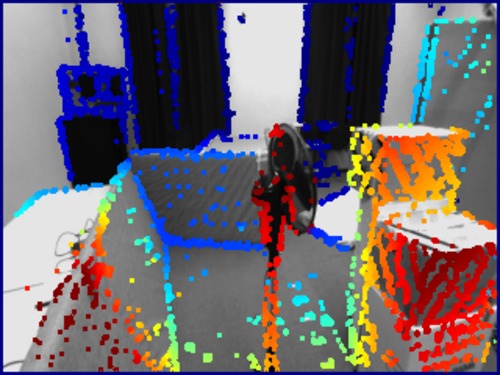}\hspace{-0.5mm}
\includegraphics[width=.163\linewidth]{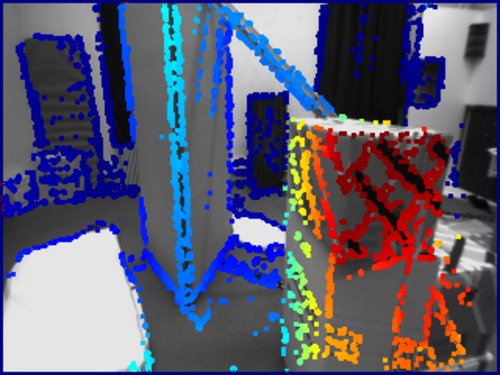}\hfill
\includegraphics[width=.163\linewidth]{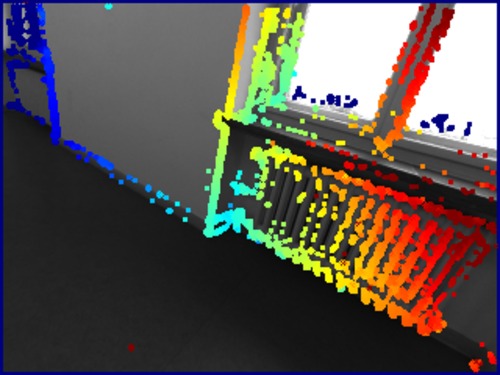}\hspace{-0.5mm}
\includegraphics[width=.163\linewidth]{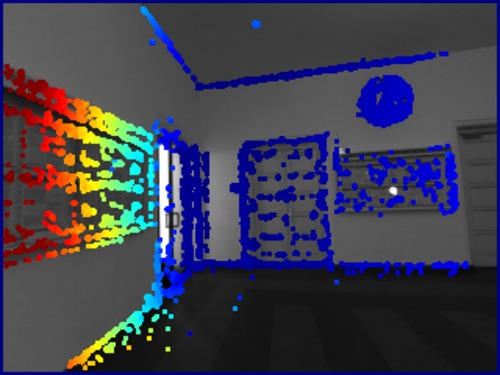}\hfill
\includegraphics[width=.163\linewidth]{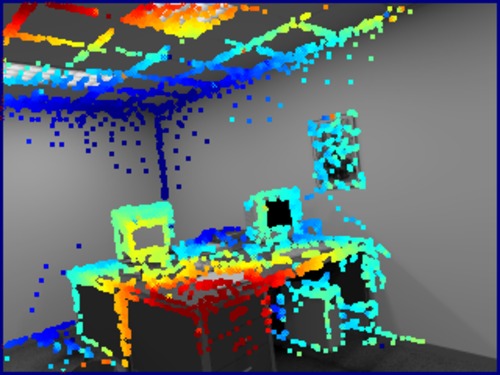}\hspace{-0.5mm}
\includegraphics[width=.163\linewidth]{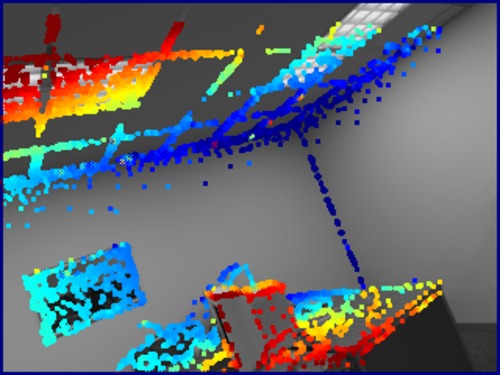}\\
\caption{\textbf{Qualitative examples.} One scene from each dataset (left to right: 
\textit{V2\_01\_easy} \cite{burrii6ijrr}, \textit{seq\_38} \cite{engel16archiveDataset} and 
\textit{office\_1} \cite{handa14icra}), computed in real time with default settings.
The bottom shows some corresponding (sparse) depth maps -- some scenes contain very little texture, making them very 
challenging for indirect approaches.}\vspace{-3mm}
\label{fig:examplesQualitative}
\end{figure*}

	Figure~\ref{fig:examplesQualitative} shows three more scenes (one from each dataset), together with some corresponding depth maps. 
	Note that our approach is able to track through scenes with very little texture, whereas indirect approaches fail.
	All reconstructions shown are simply accumulated from the odometry, without integrating loop-closures. 
	See the supplementary video for more qualitative results.

	\vspace*{-1mm}\section{Conclusion}\vspace*{-0.5mm}
	\label{secConclusion}
	We have presented a novel \textit{direct} and \textit{sparse} formulation for Structure from Motion. It combines the benefits of 
	direct methods (seamless ability to use \& reconstruct all points instead of only corners) 
	with the flexibility of sparse approaches (efficient, joint optimization of all model parameters). 
	This is possible in real time by omitting the geometric prior used by other direct methods, and instead evaluating the photometric 
	error for each point over a small neighborhood of pixels, to well-constrain the overall problem. Furthermore, we incorporate
	full photometric calibration, completing the intrinsic camera model that traditionally only reflects the geometric component of
	the image formation process.
	
	We have implemented our direct \& sparse model in the form of a monocular visual odometry algorithm (DSO), 
	incrementally marginalizing / eliminating old states to maintain real-time performance. To this end we have
	developed a front-end that performs data-selection and provides accurate initialization for optimizing the highly non-convex
	energy function.
	Our comprehensive evaluation on several hours of video shows the superiority of the presented formulation relative to
	state-of-the-art indirect methods.
	We furthermore present an exhaustive parameter study, indicating that (1) simply using more data does not increase
	tracking accuracy (although it makes the 3D models denser), (2) using all points instead of only corners does provide a 
	real gain in accuracy and robustness, and
	(3) incorporating photometric calibration does increase performance, in particular compared to the 
	basic \quotes{brightness constancy} assumption. 
	
	\hl{We have also shown experimentally that the indirect approach -- modeling a geometric error -- is much more
	robust to geometric noise, e.g., originating from a poor intrinsic camera calibration or a rolling shutter.
	The direct approach is in turn more robust to photometric noise, and achieves superior accuracy on well-calibrated data.
	We believe this to be one of the main explanations for the recent revival of direct formulations after a dominance of 
	indirect approaches for more than a decade:
	For a long time, the predominant source of digital image data were cameras, which originally were designed to
	capture images for human viewing (such as off-the-shelf webcams or integrated smartphone cameras).
	In this setting, the strong geometric distortions caused by rolling shutters and imprecise lenses favored the indirect approach.
	In turn, with 3D computer vision becoming an integral part of mass-market products (including autonomous cars 
	and drones, as well as mobile devices for VR and AR), cameras are being developed specifically for this purpose,
	featuring global shutters, precise lenses, and high frame-rates -- which allows direct formulations to realize their full potential.}

	Since the structure of the proposed direct sparse energy formulation is the same as that of indirect methods,
	it can be integrated with other optimization frameworks like (double-windowed) bundle adjustment \cite{strasdat_double_2011} or 
	incremental smoothing and mapping \cite{Kaess12ijrr}. 
	The main challenge here is the greatly increased degree of non-convexity compared to the indirect model, 
	which originates from the inclusion of the image in the error function -- this is likely to restrict the
	use of our model to video processing.
	
{\small
\bibliographystyle{ieee}
\bibliography{main}

\begin{thebibliography}{10}\itemsep=-1pt

\bibitem{burrii6ijrr}
M.~Burri, J.~Nikolic, P.~Gohl, T.~Schneider, J.~Rehder, S.~Omari, M.~Achtelik,
  and R.~Siegwart.
\newblock The {EuRoC} micro aerial vehicle datasets.
\newblock {\em International Journal of Robotics Research}, 2016.

\bibitem{caruso2015iros}
D.~Caruso, J.~Engel, and D.~Cremers.
\newblock Large-scale direct {SLAM} for omnidirectional cameras.
\newblock In {\em International Conference on Intelligent Robot Systems
  (IROS)}, 2015.

\bibitem{civera_inverse_2008}
J.~Civera, A.~Davison, and J.~Montiel.
\newblock Inverse depth parametrization for monocular {SLAM}.
\newblock {\em Transactions on Robotics}, 24(5):932--945, 2008.

\bibitem{davison07pami}
A.~Davison, I.~Reid, N.~Molton, and O.~Stasse.
\newblock {MonoSLAM}: Real-time single camera {SLAM}.
\newblock {\em Transactions on Pattern Analysis and Machine Intelligence
  (TPAMI)}, 29, 2007.

\bibitem{engel14eccv}
J.~Engel, T.~Sch\"ops, and D.~Cremers.
\newblock {LSD-SLAM}: Large-scale direct monocular {SLAM}.
\newblock In {\em European Conference on Computer Vision (ECCV)}, 2014.

\bibitem{engel15iros}
J.~Engel, J.~Stueckler, and D.~Cremers.
\newblock Large-scale direct slam with stereo cameras.
\newblock In {\em International Conference on Intelligent Robot Systems
  (IROS)}, 2015.

\bibitem{engel14ras}
J.~Engel, J.~Sturm, and D.~Cremers.
\newblock Scale-aware navigation of a low-cost quadrocopter with a monocular
  camera.
\newblock {\em Robotics and Autonomous Systems (RAS)}, 62(11):1646--–1656,
  2014.

\bibitem{engel16archiveDataset}
J.~Engel, V.~Usenko, and D.~Cremers.
\newblock A photometrically calibrated benchmark for monocular visual odometry.
\newblock In {\em arXiv preprint arXiv}, 2016.

\bibitem{forster14icra}
C.~Forster, M.~Pizzoli, and D.~Scaramuzza.
\newblock {SVO}: Fast semi-direct monocular visual odometry.
\newblock In {\em International Conference on Robotics and Automation (ICRA)},
  2014.

\bibitem{handa14icra}
A.~Handa, T.~Whelan, J.~McDonald, and A.~Davison.
\newblock A benchmark for {RGB-D} visual odometry, {3D} reconstruction and
  {SLAM}.
\newblock In {\em International Conference on Robotics and Automation (ICRA)},
  2014.

\bibitem{huang200iser}
G.~P. Huang, A.~I. Mourikis, and S.~I. Roumeliotis.
\newblock A first-estimates {J}acobian {EKF} for improving {SLAM} consistency.
\newblock In {\em International Symposium on Experimental Robotics}, 2008.

\bibitem{jin00cvpr}
H.~Jin, P.~Favaro, and S.~Soatto.
\newblock Real-time 3-d motion and structure of point features: Front-end
  system for vision-based control and interaction.
\newblock In {\em International Conference on Computer Vision and Pattern
  Recognition (CVPR)}, 2000.

\bibitem{jin03js}
H.~Jin, P.~Favaro, and S.~Soatto.
\newblock A semi-direct approach to structure from motion.
\newblock {\em The Visual Computer}, 19(6):377--394, 2003.

\bibitem{Kaess12ijrr}
M.~Kaess, H.~Johannsson, R.~Roberts, V.~Ila, J.~Leonard, and F.~Dellaert.
\newblock {iSAM2}: Incremental smoothing and mapping using the {B}ayes tree.
\newblock {\em International Journal of Robotics Research}, 31(2):217--236, Feb
  2012.

\bibitem{kerl15iccv}
C.~Kerl, J.~Stueckler, and D.~Cremers.
\newblock Dense continuous-time tracking and mapping with rolling shutter
  {RGB-D} cameras.
\newblock In {\em International Conference on Computer Vision (ICCV)}, 2015.

\bibitem{klein07ismar}
G.~Klein and D.~Murray.
\newblock Parallel tracking and mapping for small {AR} workspaces.
\newblock In {\em International Symposium on Mixed and Augmented Reality
  (ISMAR)}, 2007.

\bibitem{leutenegger2015ijrr}
S.~Leutenegger, S.~Lynen, M.~Bosse, R.~Siegwart, and P.~Furgale.
\newblock Keyframe-based visual--inertial odometry using nonlinear
  optimization.
\newblock {\em International Journal of Robotics Research}, 34(3):314--334,
  2015.

\bibitem{li2013icra}
M.~Li, B.~Kim, and A.~Mourikis.
\newblock Real-time motion estimation on a cellphone using inertial sensing and
  a rolling-shutter camera.
\newblock In {\em International Conference on Robotics and Automation (ICRA)},
  2013.

\bibitem{lovegrove2013spline}
S.~Lovegrove, A.~Patron-Perez, and G.~Sibley.
\newblock Spline fusion: A continuous-time representation for visual-inertial
  fusion with application to rolling shutter cameras.
\newblock In {\em British Machine Vision Converence (BMVC)}, 2013.

\bibitem{mur2015orb}
R.~Mur-Artal, J.~Montiel, and J.~Tardos.
\newblock {ORB-SLAM}: a versatile and accurate monocular {SLAM} system.
\newblock {\em Transactions on Robotics}, 31(5):1147--1163, 2015.

\bibitem{newcombe2011iccv}
R.~Newcombe, S.~Lovegrove, and A.~Davison.
\newblock {DTAM}: Dense tracking and mapping in real-time.
\newblock In {\em International Conference on Computer Vision (ICCV)}, 2011.

\bibitem{pizzoli14icra}
M.~Pizzoli, C.~Forster, and D.~Scaramuzza.
\newblock {REMODE}: Probabilistic, monocular dense reconstruction in real time.
\newblock In {\em International Conference on Robotics and Automation (ICRA)},
  2014.

\bibitem{ranftl16cvpr}
R.~Ranftl, V.~Vineet, Q.~Chen, and V.~Koltun.
\newblock Dense monocular depth estimation in complex dynamic scenes.
\newblock In {\em International Conference on Computer Vision and Pattern
  Recognition (CVPR)}, 2016.

\bibitem{schoeps14ismar}
T.~Sch\"{o}ps, J.~Engel, and D.~Cremers.
\newblock Semi-dense visual odometry for {AR} on a smartphone.
\newblock In {\em International Symposium on Mixed and Augmented Reality
  (ISMAR)}, 2014.

\bibitem{strasdat_double_2011}
H.~Strasdat, A.~J. Davison, J.~M.~M. Montiel, and K.~Konolige.
\newblock Double window optimisation for constant time visual {SLAM}.
\newblock In {\em International Conference on Computer Vision (ICCV)}, 2011.

\bibitem{stuehmer10dagm}
J.~St\"uhmer, S.~Gumhold, and D.~Cremers.
\newblock Real-time dense geometry from a handheld camera.
\newblock In {\em Pattern Recognition (DAGM)}, 2010.

\bibitem{valgaerts2012dense}
L.~Valgaerts, A.~Bruhn, M.~Mainberger, and J.~Weickert.
\newblock Dense versus sparse approaches for estimating the fundamental matrix.
\newblock {\em International Journal of Computer Vision (IJCV)},
  96(2):212--234, 2012.

\end{thebibliography}
}

\end{document}